%% file: main.tex
\documentclass[10pt,twocolumn,letterpaper]{article}

\usepackage{cvpr}              

\input{preamble}

\definecolor{cvprblue}{rgb}{0.21,0.49,0.74}
\usepackage[pagebackref,breaklinks,colorlinks,allcolors=cvprblue]{hyperref}

\def\paperID{} 
\def\confName{CVPR}
\def\confYear{2026}

\title{3D-IDE: 3D Implicit Depth Emergent}

\author{
Chushan Zhang\textsuperscript{1,3} \quad 
Ruihan Lu\textsuperscript{2,3} \quad 
Jinguang Tong\textsuperscript{1} \quad 
Yikai Wang\textsuperscript{4$\dagger$} \quad 
Hongdong Li\textsuperscript{1$\dagger$} \\[2ex]
{\small
\begin{tabular}{cc}
\textsuperscript{1}School of Computing, Australian National University & \textsuperscript{2}School of EECS, The University of Queensland \\
\textsuperscript{3}FreiNexus & \textsuperscript{4}School of Artificial Intelligence, Beijing Normal University
\end{tabular}
}
\\[1ex] 
{\small $\dagger$ Corresponding authors} 
\vspace{-1em}
}

\begin{document}
\maketitle
\input{sec/0_abstract}

\input{sec/1_intro}

\input{sec/2_related}

\input{sec/3_method}

\input{sec/4_experiment}

\input{sec/5_conclusion}


\clearpage

\noindent \textbf{Acknowledgement}: YW is supported by the NSFC (No.~62576043). HL is supported in part by an ARC Grant (DP220100800). HL holds concurrent appointments with both ANU and Amazon; however, this research was conducted at ANU and is independent of Amazon.   
{
    \small
    \bibliographystyle{ieeenat_fullname}
    \bibliography{main}
}

\input{sec/X_suppl}
\end{document}

%% file: preamble.tex
\newcounter{commentctr}

\usepackage{booktabs}
\usepackage{multirow}
\usepackage{threeparttable}
\usepackage[table]{xcolor} 
\definecolor{myblue}{RGB}{230,245,255}
\usepackage{svg}
\usepackage{tikz}
\usepackage{pifont}
\usetikzlibrary{external}
\usepackage{booktabs}
\usepackage{tabularx}
\usepackage[table]{xcolor}  
\usepackage{siunitx}

\newcommand{\smallnabla}{\scalebox{0.75}{$\nabla$}}

\usepackage{booktabs}
\usepackage[table]{xcolor} 
\usepackage{tabularx}
\definecolor{Gray}{gray}{0.90}
\definecolor{Orange}{rgb}{1,0.8,0.5}
\usepackage{array}

\usepackage[normalem]{ulem}

%% file: sec/0_abstract.tex
\begin{abstract}
Leveraging 3D information within Multimodal Large Language Models (MLLMs) has recently shown significant advantages for indoor scene understanding. However, existing methods, including those using explicit ground-truth 3D positional encoding and those grafting external 3D foundation models for implicit geometry, struggle with the trade-off in 2D-3D representation fusion, leading to suboptimal deployment. To this end, we propose 3D-Implicit Depth Emergence, a method that reframes 3D perception as an emergent property derived from geometric self-supervision rather than explicit encoding. Our core insight is the Implicit Geometric Emergence Principle: by strategically leveraging privileged geometric supervision through mechanisms like a fine-grained geometry validator and global representation constraints, we construct an information bottleneck. This bottleneck forces the model to maximize the mutual information between visual features and 3D structures, allowing 3D awareness to emerge naturally within a unified visual representation. Unlike existing approaches, our method enables 3D perception to emerge implicitly, disentangling features in dense regions and, crucially, eliminating depth and pose dependencies during inference with zero latency overhead. This paradigm shift from external grafting to implicit emergence represents a fundamental rethinking of 3D knowledge integration in visual-language models. Extensive experiments demonstrate that our method surpasses SOTA on multiple 3D scene understanding benchmarks. Our approach achieves a 55\% reduction in inference latency while maintaining strong performance across diverse downstream tasks, underscoring the effectiveness of meticulously designed auxiliary objectives for dependency-free 3D understanding. Source code can be found at \href{https://chushanzhang.github.io/3D-IDE/}{\textcolor[HTML]{E8477A}{\small\texttt{github.com/ChushanZhang/3D-IDE}}}.


\vspace{-1em}
\end{abstract}

%% file: sec/1_intro.tex
\begin{figure}[ht!] 
\centering 
\includegraphics[width=0.5\textwidth]{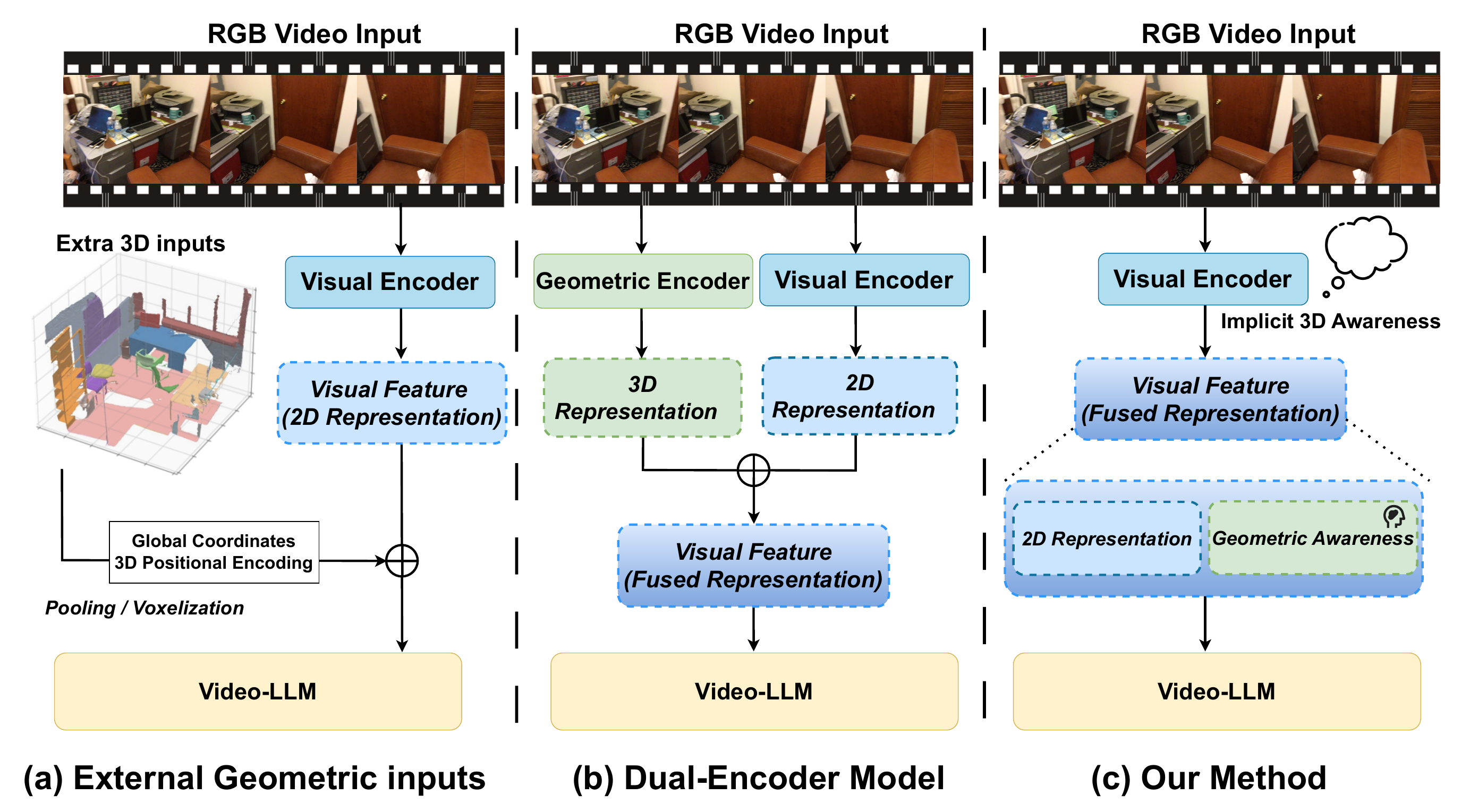}
\caption{\textbf{Comparison of 3D-aware designs for video-LLMs.}
\textbf{(a)} Explicit coordinate injection fuses 2D features with coarse 3D positional embeddings and requires 3D inputs at inference. \textbf{(b)} Dual encoders separately process RGB and geometry, then fuse their outputs, increasing complexity and latency. \textbf{(c)} 3D-IDE uses a single visual encoder trained so that 3D awareness emerges implicitly, enabling efficient RGB-only inference.
}
\vspace{-1.5em}
\label{fig:framework_comparison}
\end{figure}

\section{Introduction}
\label{sec:intro}
Multimodal large language models (MLLMs) have rapidly become a unifying interface for 3D scene understanding, thanks to their strong 2D priors and powerful reasoning ability~\cite{llava,llava-1.5,gpt-4}. Recent work adapts MLLMs to 3D through several routes: point cloud encoders that project 3D features into language-aligned spaces~\cite{pointllm,ll3da}, multi-view approaches that lift 2D features to 3D representations~\cite{3D-LLM_huang,llava-3d}, and video-based methods that capture spatio-temporal relationships in 3D scenes~\cite{video-llm,liu2024oryx,DGNS}. Among these paradigms, video-based MLLMs are especially attractive because they naturally preserve scene continuity and leverage pretrained video understanding capabilities.

Within this video-based paradigm, existing methods mainly differ in how they inject geometric awareness. As illustrated in~\cref{fig:framework_comparison}, one line of work explicitly encodes per-pixel 3D coordinates from depth and camera poses into visual tokens, requiring additional 3D inputs such as depth maps or camera poses at inference time~\cite{video-llm,3drs,rgbd-cons}. Another line relies on a separate 3D foundation model whose features are fused with those of a 2D visual encoder, forming a dual-encoder design~\cite{vid-llm,injectvggt,flare,vggt}. Both strategies, however, have fundamental drawbacks. Explicit coordinate injection introduces a critical dependency on specialized 3D sensors and passes through aggressive downsampling and quantization, which collapses fine-grained geometry and lead to a ``double information loss''. Dual-encoder designs increase parameters and latency and create a representation gap: 2D and 3D encoders are typically frozen and optimized under different objectives, so the language model is forced to act as a late-stage aligner instead of focusing on high-level 3D reasoning, ultimately limiting deployability.

This work asks a different question: can we learn a 3D-aware representation that uses only RGB video at inference, yet retains geometry strong enough for 3D grounding and reasoning? To answer this, we propose \textbf{3D-IDE} (3D-Implicit Depth Emergence), guided by the \textbf{Implicit Geometric Emergence Principle (IGEP)}. Rather than treating geometry as a mandatory input, we regard it as privileged supervision that is available only during training. A lightweight, training-only geometric validator and a global 3D teacher provide fine-grained and scene-level geometric signals that push the visual encoder to embed 3D structure directly in its tokens, without modifying the inference-time interface or introducing any additional inputs.

Concretely, the same visual tokens that condition the video-LLM are also required, during training, to support dense, uncertainty-aware depth prediction, localized cross-view consistency across neighboring frames, and alignment with a frozen 3D foundation model~\cite{flare,vggt}. Because the geometric head is deliberately low-capacity and discarded at test time, the model cannot rely on it as a 3D expert; instead, it must internalize geometric cues in the shared encoder. This implicit training pressure yields a single RGB-only representation that is geometrically informative, avoids explicit coordinate injection and separate 3D encoders, and adds no latency or extra inputs at deployment. Overall our main contributions are three-fold:

\begin{itemize}
    \item We introduce the Implicit Geometric Emergence Principle, which views 3D awareness in video-MLLMs as an emergent effect of training-time geometric supervision rather than explicit 3D inputs or heavy 3D encoders.
    \item We realize this principle in \textbf{3D-IDE}, which uses a lightweight geometric validator and a global 3D teacher to impose uncertainty-aware depth, multi-view consistency, and scene-level constraints, while keeping inference strictly RGB-only.
    \item On standard 3D grounding, captioning, and QA benchmarks, 3D-IDE outperforms prior RGB-only video-MLLMs and remains competitive with methods using explicit 3D inputs, achieving up to $6.36\%$ higher 3D grounding accuracy with $12.86\%$ fewer parameters and $55.28\%$ faster inference.
\end{itemize}

%% file: sec/2_related.tex
\section{Related Work}
\label{sec:related}

\paragraph{MLLMs for 3D Scene Understanding.} 
Adapting Multimodal Large Language Models (MLLMs)~\cite{llava, qwen2, llava-onevision, gpt-4} for 3D scene understanding has attracted significant interest. Early pioneering works focused on bridging the modality gap between 3D representations and language. These methods typically ingest point cloud data, which is processed by a specialized 3D encoder (like PointNet~\cite{pointnet} or its variants~\cite{pointnet++}) before being projected into the MLLM's embedding space. Prominent examples in this category include PointLLM~\cite{Xu2024PointLLM}, 3D-LLM~\cite{3D-LLM_huang}, Chat-3D~\cite{Wang2023Chat3D}, LL3DA~\cite{Chen2024LL3DA}, and Grounded 3D-LLM~\cite{Chen2024Grounded3DLLM}. While effective for 3D-centric tasks, these approaches face two key challenges: the scarcity of large-scale, well-annotated 3D–text datasets and a fundamental disconnect from the rich 2D visual knowledge learned by MLLMs during large-scale pre-training. Recent empirical studies~\cite{el2024probing, man2024lexicon3d} have shown that 2D pre-trained visual foundation models can effectively extract 3D spatial representations from 2D features, indicating that large-scale 2D pre-training inherently encodes structural priors of 3D scenes. Building on this insight, recent work has shifted toward video-based approaches~\cite{video-llm, qi2025gpt4scene}, which process 3D scenes as multi-view sequences or video frames to naturally preserve scene continuity, exploit pre-trained video understanding capabilities, and capture spatio-temporal relationships across frames.

\vspace{-1em}

\paragraph{3D-Aware Integration in Video-MLLMs.} Within the video or multi-view paradigm, contemporary work incorporates 3D priors into MLLMs along two explicit routes and one implicit route. First, direct injection methods, exemplified by Video-3D LLM~\cite{video-llm} and its variant 3DRS~\cite{3drs}, treat 3D scenes as videos and augment patch-level visual tokens with global 3D coordinates computed from depth and camera poses, a strategy that is also adopted by LLaVA-3D~\cite{llava-3d} and GPT4Scene~\cite{qi2025gpt4scene}. Second, explicit supervision or fusion methods refine these representations through geometric feature fusion: Vid-LLM~\cite{vid-llm} aligns MLLM visual features to those from pre-trained 3D foundation models such as VGGT~\cite{vggt} and FLARE~\cite{flare}, while VG-LLM~\cite{injectvggt} extracts priors such as inter-frame correspondences from RGB videos using a pre-trained 3D geometry encoder and fuses them with 2D tokens; Geometry supervision is also used in 3D reconstruction \cite{2DGS}. In contrast, 3D-IDE advances an implicit route via the Implicit Geometric Emergence Principle, where geometry serves as a form of privileged training signal for an auxiliary validator, encouraging the MLLMs to internally emerge 3D-aware structure directly from monocular cues without relying on explicit 3D priors or any additional inference-time components.

%% file: sec/3_method.tex
\section{Method}
\label{sec:method}
This section details our proposed 3D-IDE framework, which is designed to overcome the critical limitations of existing MLLMs in 3D scene understanding. In \cref{subsec:limitation}, we begin by formally revisiting and analyzing the distinct structural constraints of the two conventional paradigms, which motivates our approach. We then introduce the key principles of our proposed 3D-IDE framework in \cref{subsec:ours method}, centered on the Implicit Geometric Emergence Principle. Following this, we detail the specific architecture and composite training objectives used to realize this principle in \cref{subsec:ours training}. The overall framework is illustrated in \cref{fig:framework_overall}.

\subsection{Preliminaries}

\label{subsec:limitation}
We first formalize the typical structure of 3D-aware video MLLMs and the two dominant paradigms used to inject geometric information.
Let $\{I_t\}_{t=1}^N$ denote a set of multi-view images or video frames, and let $f_E$ be a visual encoder. For each frame $I_t$, the encoder produces patch-level features
\[
F_t = f_E(I_t) \in \mathbb{R}^{H' \times W' \times d},
\]
where $H',W'$ are the downsampled spatial dimensions and $d$ is the feature dimension. 
Existing 3D-aware designs construct enriched features $F_t^{3D}$ by combining $F_t$ with 3D cues, typically from explicit inputs or external encoders.

\input{tabs/ablation_gt_input}

\begin{figure}[th!]
    \centering
    \begin{subfigure}[b]{0.33\columnwidth} 
        \includegraphics[width=\linewidth]{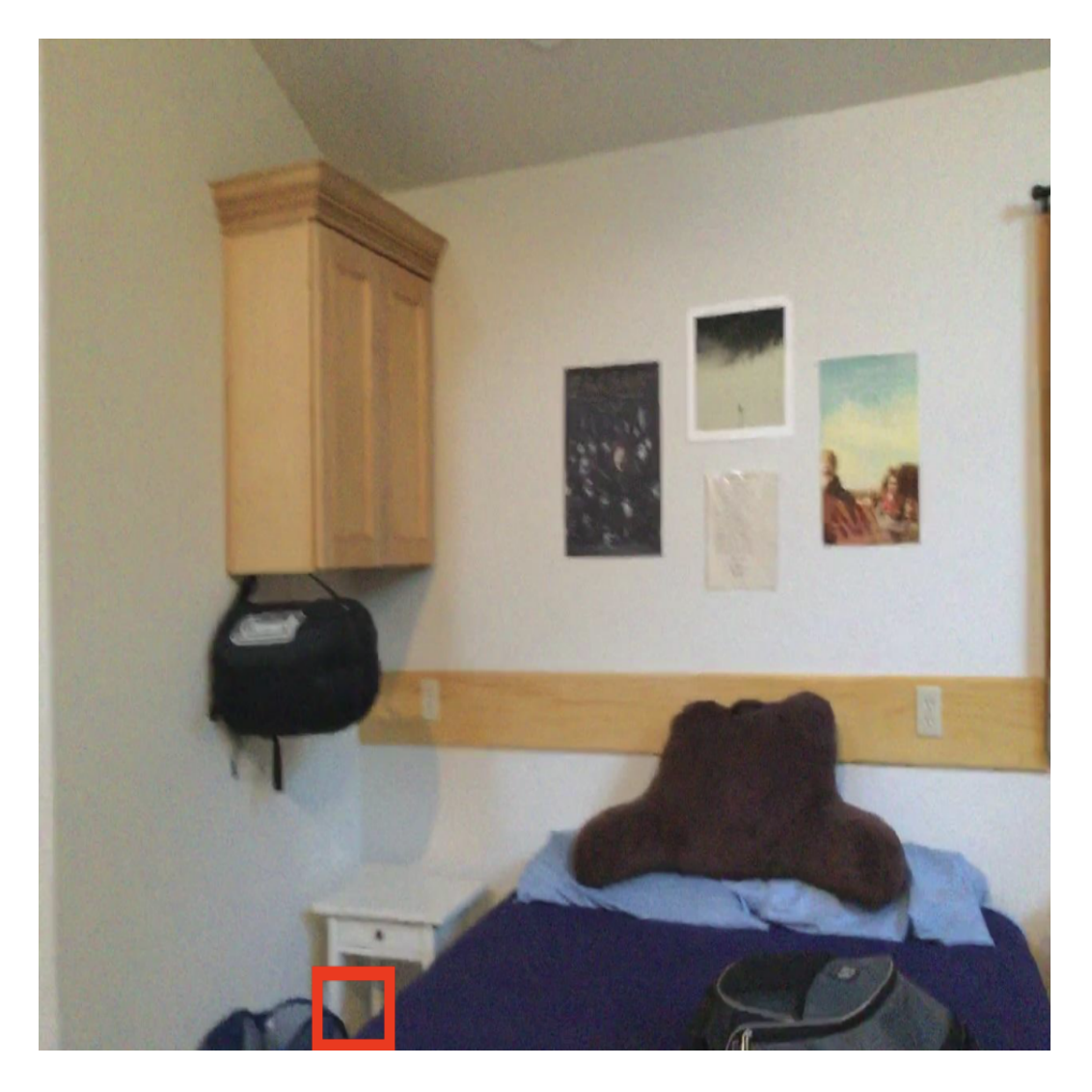} 
        \caption{RGB input}
        \label{fig:sub1}
    \end{subfigure}%
    \hfill%
    \begin{subfigure}[b]{0.35\columnwidth} 
        \includegraphics[width=\linewidth]{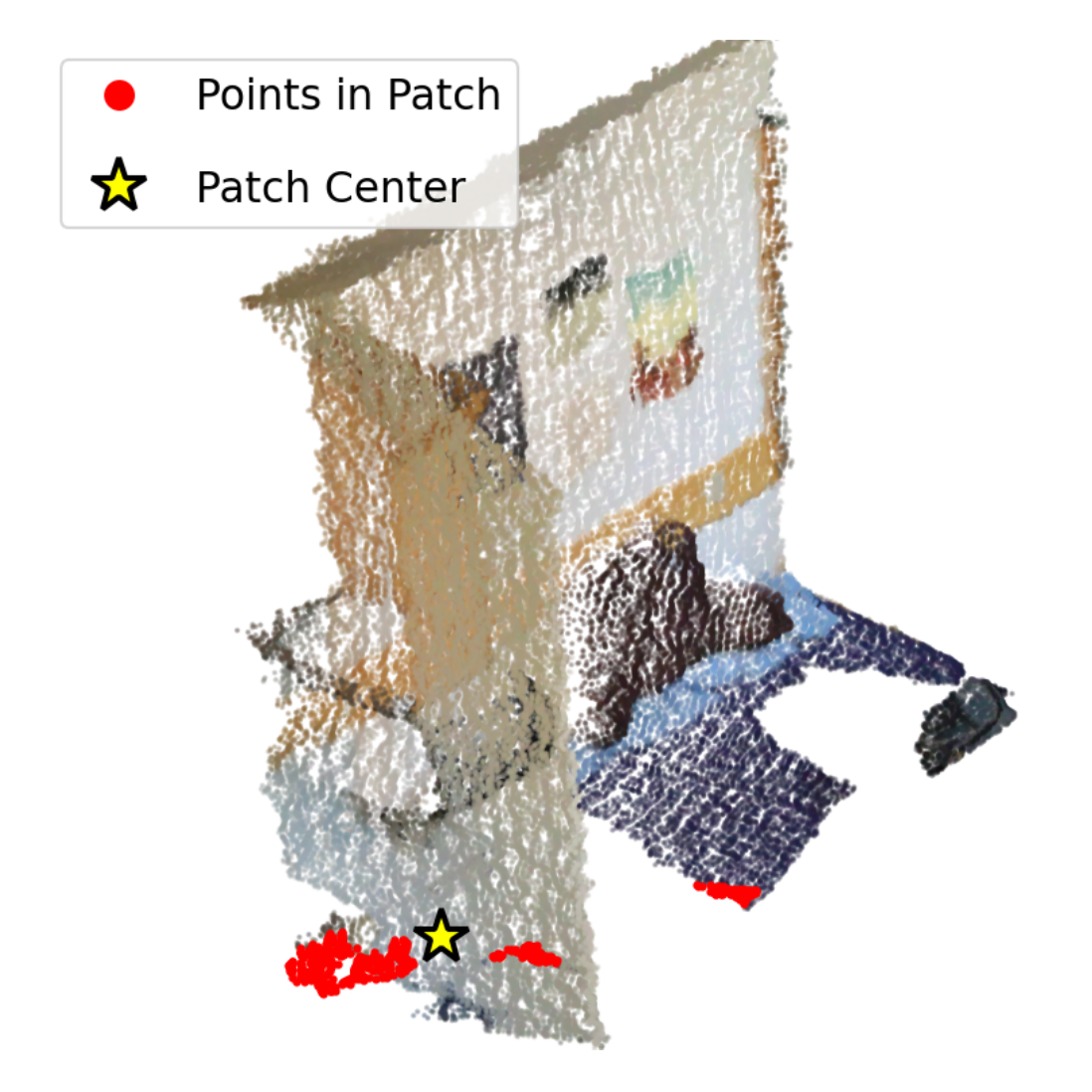} 
        \caption{Pooling loss}
        \label{fig:sub2}
    \end{subfigure}%
    \hfill%
    \begin{subfigure}[b]{0.31\columnwidth} 
        \includegraphics[width=\linewidth]{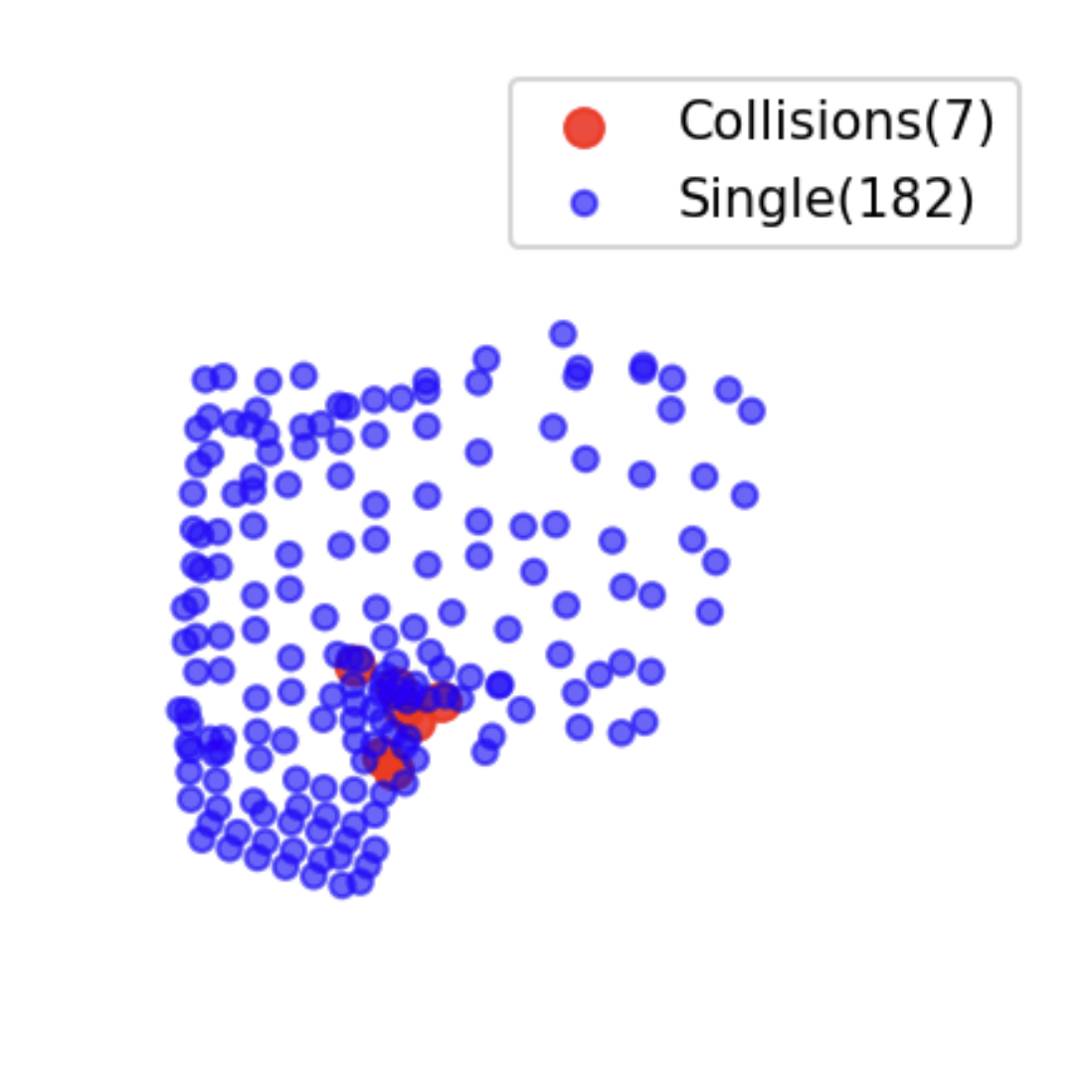} 
        \caption{Voxelization loss}
        \label{fig:sub3}
    \end{subfigure}
     \caption{
     Illustration of the double information loss in explicit coordinate injection. \textbf{(a)} RGB frame with a 2D patch whose pixels are back-projected to point cloud. \textbf{(b)} Pooling collapses all patch points into one token, losing local structure. \textbf{(c)} Voxelization merges distinct 3D points into the same voxel, further degrading fine-grained geometry and harming downstream 3D reasoning.
    }
    \label{fig:double_information loss}
    \vspace{-11pt}
\end{figure}

\begin{figure*}[t]
    \centering
    \includegraphics[width=1.05 \linewidth]{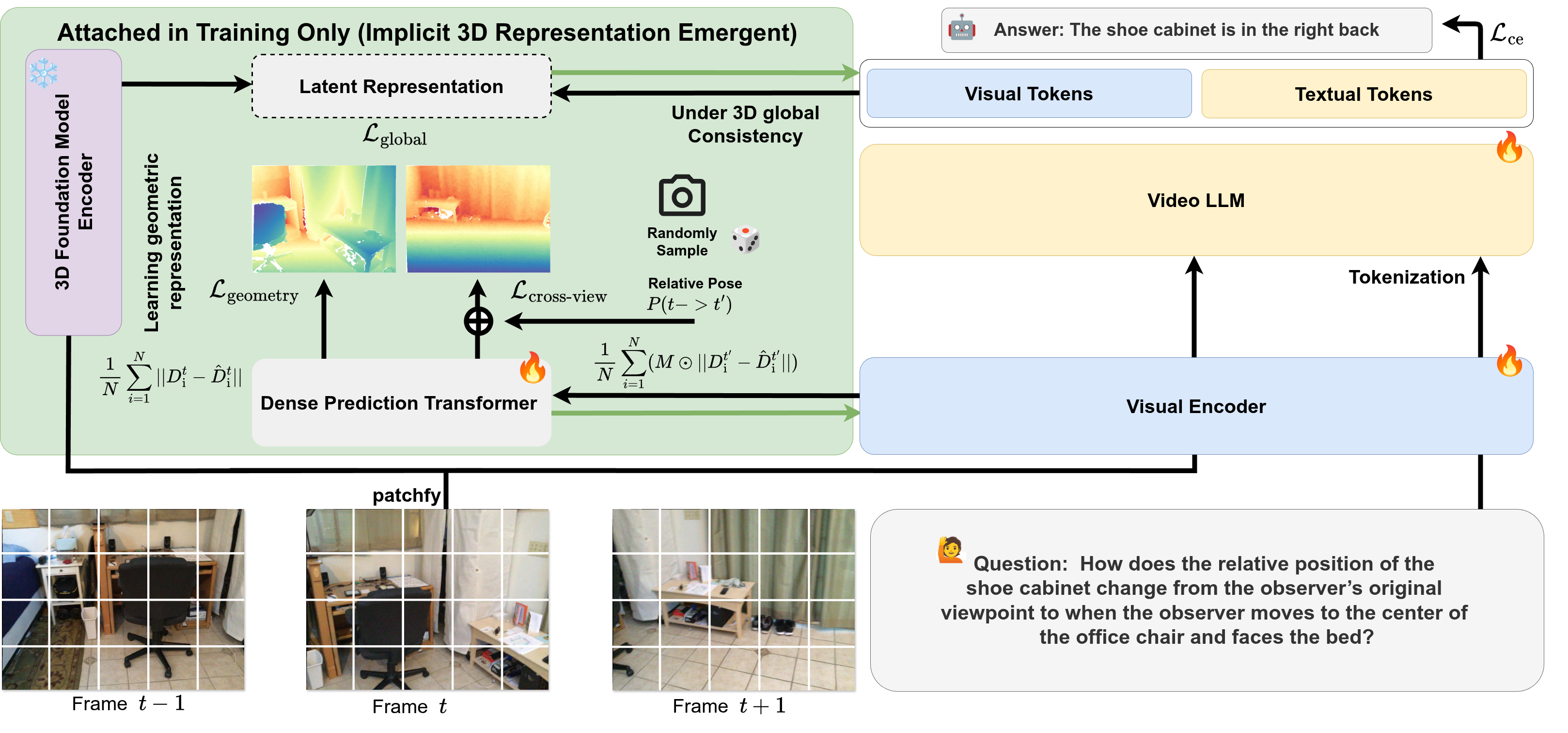} 
    \caption{ \textbf{The 3D-IDE framework}. Our approach avoids the ``Double Information Loss" (see \cref{fig:double_information loss}) inherent in explicit coordinate injection methods. Instead of injecting coarse, lossy coordinates, we use a privileged training module (green box) that is detached at inference for \textbf{zero latency}. This module forces the model to learn a fine-grained 3D representation implicitly via two parallel gradient signals (green arrows): a geometric gradient from a weak depth validator and a global gradient from a frozen foundation model guidance.} 
    \label{fig:framework_overall} 
    \vspace{-1em}
\end{figure*}

\paragraph{Explicit 3D-Input.} 
Methods in this family rely on explicit 3D inputs, such as depth maps. This reliance is a critical flaw, significantly limiting generalization and practical deployment in real-world scenarios where such data is unavailable. To quantify this dependency, we ablate the 3D ground-truth inputs during inference. As shown in \cref{tab:remove3D-performance}, we observe a substantial performance degradation when 3D inputs are withheld. Despite being fine-tuned on 3D data, the model's performance collapses to a level comparable to a 2D MLLM (1st row) in a zero-shot setting. This confirms that existing models often use 3D inputs as a crutch, failing to develop robust, generalizable 3D understanding. This method, used by \cite{video-llm, 3drs}, requires a per-pixel 3D coordinate map $C_t \in \mathbb{R}^{H \times W \times 3}$ associated with each image $I_t$, computed by back-projecting the depth map using camera parameters. These coordinates are encoded via a positional encoding function $\phi(\cdot)$ and injected into the visual features:
\begin{equation}
\label{eq:method1}
F^{3D}_t = F_t + \phi(C_t)
\end{equation}
This encoding process itself is highly problematic. As visualized in \cref{fig:double_information loss}, the coordinates $C_t$ are heavily downsampled and voxelized to align with patch-wise features. This coarse aggregation creates an information bottleneck, causing fine-grained geometric structures to be averaged out and rendered indistinguishable. This double information loss prevents the model from perceiving fine-grained geometry and exacerbates spatial ambiguity, especially when a single patch is geometrically complex or contains multiple objects.

\vspace{-0.5em}
\paragraph{External 3D-Encoder.} 
A second family of approaches employs a dedicated geometric encoder $E_{\text{geo}}$ to extract latent 3D features from the input frames. Given image features $F_t$ from the visual encoder and geometric features $G_t = E_{\text{geo}}(I_t)$, the 3D-aware representation is obtained by a feature fusion $\oplus$, which is formulated as follows:
\begin{equation}
\label{eq:method2}
F^{3D}_t = F_t \oplus G_t,
\end{equation}
Such geometric encoders are often large models (e.g., VGGT~\cite{vggt} has on the order of one billion parameters), which substantially increases the overall model size and inference cost and makes end-to-end optimization more demanding. In addition, $E_{\text{geo}}$ is typically pre-trained and kept frozen under objectives different from those of the video MLLM, so the resulting feature spaces may be poorly aligned. As a consequence, the language model must implicitly learn to reconcile the 2D and 3D streams, limiting the effectiveness of late fusion for 3D reasoning.

\vspace{-0.5em}

\paragraph{LLMs with 3D visual tokens.}
Regardless of the method used to obtain $F^{3D}_t$ (Eq.~(\ref{eq:method1}) or (\ref{eq:method2})), given a tokenized text instruction $\psi$, models in this dependency paradigm are optimized by minimizing the standard cross-entropy loss:
\begin{equation}
\label{eq:loss_ce}
    \mathcal{L}_{ce} = -\sum_{n=1}^{T} \log p_\theta(y_n | y_{<n}, \{F^{3D}_t\}_{t=1}^N, \psi),
\end{equation}
where $y_n$ is the $n$-th ground-truth output token, $T$ is the total length of the target sequence, and $\theta$ denotes the set of all trainable parameters of the entire MLLM model $f_\theta$. The analyses above highlight a critical trilemma in 3D-aware MLLMs: existing methods either (1) rely on unavailable ground-truth inputs, (2) destroy geometric information via coarse encoding, or (3) depend on massive, external 3D foundation models. These limitations motivate the need for a 3D representation that is simultaneously lightweight, information-preserving, and independent of ground-truth.


\subsection{Implicit Geometric Emergence Principle}
\label{subsec:ours method}
To address the trilemma of inference latency, external dependency, and representation misalignment, we propose the Implicit Geometric Emergence Principle. The core idea is to reframe 3D perception as an emergent property derived from privileged supervision rather than external inputs. 

\subsubsection{Geometric Emergence Learning}
Building on the previous analysis, our objective is to obtain a 3D-aware representation that requires only RGB inputs at inference, yet retains sufficient geometric information for downstream reasoning. In 3D-IDE, we treat 3D awareness as an emergent property of the encoder features themselves. Formally, within our unified representation learning framework, we redefine the 3D-aware feature as:

\begin{equation}
\label{eq:ours-feature}
F_{t}^{3D} \equiv F_{t},
\end{equation}

so that a single feature space must encode both 2D semantics and 3D geometry, rather than being an explicit concatenation or addition of separate 2D and 3D streams.

To induce such an implicit 3D representation, we regard geometry as privileged supervision that is only available during training. Let $Y^{3D}$ denote latent 3D scene variables associated with the input sequence, such as depth, local surface structure, and multi-view consistency. Rather than providing $Y^{3D}$ (or its substitutes) as explicit input to the MLLM, we encourage the encoder to learn features from which $Y^{3D}$ can be recovered by low-capacity decoders attached only during training. This encourages the encoder to internalize geometric information while keeping the inference-time interface unmodified. Concretely, 3D-IDE attaches a lightweight, training-only privileged module to the encoder. Given the sequence of tokens $F_t$, an Auxiliary Geometric Validator $f_P$, implemented as a compact DPT-style decoder, predicts a dense depth map $\hat{D}_t$ and a pixel-wise uncertainty map $\hat{\Sigma}_t$ for each frame. A geometric supervision signal, denoted by $\mathcal{L}_{\text{geometry}}$, measures the discrepancy between $(\hat{D}_t, \hat{\Sigma}_t)$ and the available ground-truth depth, and is designed to encourage fine-grained and uncertainty-aware depth prediction.

The video setting provides additional structure through multi-view geometry. We exploit this structure in a localized and a global manner. At the local level, a cross-view term, denoted $\mathcal{L}_{\text{cross-view}}$, enforces consistency between depth predictions across neighboring frames under the known relative pose by warping and comparing outputs. This encourages the model to learn viewpoint-consistent geometry from sparsely sampled frame pairs, without incurring the cost of exhaustive pairwise supervision. However, enforcing cross-view constraints densely over all frame pairs is computationally prohibitive for long video sequences. To extend consistency beyond the sampled pairs in a lightweight manner, we therefore introduce a complementary global constraint, denoted $\mathcal{L}_{\text{global}}$. Conceptually, this constraint provides a compact scene-level regularization signal: it encourages the model to produce a global representation of the scene that is geometrically coherent, thereby propagating consistency across the entire sequence while maintaining low computational overhead.

\subsubsection{Geometric Validator Capacity and Initialization}
\label{subsubsec:weaker_better}

IGEP is designed so that most of the 3D reasoning capacity resides in the shared visual encoder. This makes the Auxiliary Geometric Validator $f_P$ conceptually a readout module rather than a task-specific 3D expert. In 3D-IDE we therefore instantiate $f_P$ as a low-capacity decoder trained from scratch, instead of a powerful pre-trained depth network.

Let $\mathcal{P}_{\text{weak}}$ denote a family of lightweight validators parameterized by $\theta_P$. Given an encoder $f_E(\cdot;\theta_E)$, the auxiliary geometric supervision can be abstracted as
\[
\min_{\theta_E, \theta_P} \;
\mathcal{L}_{\text{geometry}}(f_P(f_E(I; \theta_E); \theta_P),D^{\text{gt}}),
\;  
f_P \in \mathcal{P}_{\text{weak}}
\]



Because $f_P$ is restricted to $\mathcal{P}_{\text{weak}}$ and has no pre-training, it cannot absorb arbitrarily complex 3D reasoning. The geometric loss instead constrains the joint system $(f_E, f_P)$ and biases the optimization toward encoder representations $F_t = f_E(I_t; \theta_E)$ in which geometry is organized in a form that such a simple validator can decode. In contrast, a high-capacity, pre-trained validator would carry strong task-specific priors and could account for much of the 3D reasoning on its own, weakening the pressure on the encoder. From the perspective of IGEP, we therefore favor a weak, from-scratch validator that allocates capacity and prior knowledge primarily to the shared encoder.

Our ablation in Tab.~\ref{tab:ablation-table1} compares two instantiations of $f_P$ that share the same architecture but differ in initialization: a pre-trained depth model from vggt and a from-scratch variant. The two settings achieve very similar performance across ScanRefer and Multi3DRef, with the from-scratch variant being slightly but consistently higher. This suggests that a strong pre-trained depth head is not necessary for IGEP to be effective, and that a weak, from-scratch validator is sufficient while better matching the desired design philosophy of keeping geometry inside the shared encoder.

\subsection{Architecture and Training Objective}
\label{subsec:ours training}
Our implementation adapts a standard video-language architecture to incorporate our privileged learning framework. The core idea is a training-only module that enables the model to develop an implicit perception of 3D geometry. This module is entirely detached and discarded during inference, thus incurring zero latency overhead. The training is driven by a composite objective that primarily focuses on learning fine-grained geometry through our IGEP, while localized and global signals enforce multi-view consistency.


\paragraph{3D-Aware Visual Encoder.}
We employ a pretrained Vision Transformer backbone, SigLIP~\cite{siglip}, which is fine-tuned end-to-end. Given a sequence of $N$ video frames $\{I_t\}_{t=1}^{N}$, the encoder produces per-frame token features
\[
F_t = f_E(I_t; \theta_E) \in \mathbb{R}^{M \times C}, \quad t = 1,\dots,N,
\]
where $M$ is the number of visual tokens and $C$ is the channel dimension. As defined in Eq.~\eqref{eq:ours-feature}, these tokens are directly treated as the 3D-aware features $F_t^{3D}$. After a linear projection into the language embedding space, the sequence $\{F_t^{3D}\}_{t=1}^{N}$ is used as a soft visual prompt.

\paragraph{Auxiliary Geometric Validator.}
To expose the encoder to fine-grained geometric supervision, we attach an Auxiliary Geometric Validator $f_P$ to the visual tokens. For each frame $t$, the validator takes $F_t$ as input and predicts a dense depth map and a pixel-wise uncertainty map,
\[
(\hat{D}_t, \hat{\Sigma}_{D,t}) = f_P(F_t; \theta_P),
\]
where $\hat{D}_t \in \mathbb{R}^{H \times W}$ and $\hat{\Sigma}_{D,t} \in \mathbb{R}^{H \times W}$ are aligned with the image resolution. The validator is implemented as a DPT-style decoder with limited capacity (train from strath). It is used only during training and removed at inference.

\paragraph{Video MLLM.}
We adopt a LLaVA-Next-Video~\cite{llava-next-video} style architecture with Qwen2-7B~\cite{qwen2} as the language backbone. The projected visual tokens derived from $\{F_t^{3D}\}$ are concatenated with textual tokens and fed into the LLM as a soft visual prompt. The language model remains standard: given a tokenized instruction $\psi$ and target token sequence $\{y_n\}_{n=1}^{T}$, it is trained autoregressively with a cross-entropy objective, as detailed in~\cref{eq:loss_ce}.



\paragraph{Geometric Objective.}
Let $D^{\text{gt}}_t$ denote the ground-truth depth map for frame $t$, and let $\Omega$ be the set of valid pixels across all frames. For each valid pixel $p \in \Omega$, we define a per-pixel loss $\ell_p$ that combines data fidelity, gradient consistency, and uncertainty regularization:

\begin{equation}
\begin{aligned}
\ell_p = & \| \hat{\Sigma}_{D,p} \odot (\hat{D}_p - D^{\text{gt}}_p) \| \\
         & + \| \hat{\Sigma}_{D,p} \odot ({\smallnabla} \hat{D}_p - {\smallnabla} D^{\text{gt}}_p)\| \\
         & - \alpha \log \hat{\Sigma}_{D,p}
\end{aligned}
\end{equation}
where $\nabla$ denotes the spatial gradient operator, $\odot$ is the element-wise product, and $\alpha$ is a weighting hyperparameter. The geometric loss is the average of this per-pixel loss:
\begin{equation}
\mathcal{L}_{\text{geometry}}
=
\frac{1}{|\Omega|}
\sum_{p \in \Omega} \ell_p
\label{eq:geom_loss}
\end{equation}
\vspace{-0.1em}
This loss encourages accurate and uncertainty-aware depth reconstruction and compels the encoder features $F_t$ to carry fine-grained geometric information.

\paragraph{Localized Cross-View Consistency.}
To exploit temporal structure, we introduce a cross-view loss that enforces depth consistency between neighboring frames. For a reference frame $t$, we randomly sample a neighboring frame $t'$ and use the known relative pose $P(t' \rightarrow t)$ to warp the predicted depth $\hat{D}_{t'}$ into the viewpoint of frame $t$, obtaining $\hat{D}_{t' \rightarrow t}$. Let $M_{t' \rightarrow t} \in \{0,1\}^{H \times W}$ be a mask indicating valid warps, and let $\Omega_{t' \rightarrow t}$ be the set of pixels where $M_{t' \rightarrow t,p} = 1$. The cross-view consistency loss is:
\begin{equation}
\mathcal{L}_{\text{cross-view}}
=
\frac{1}{|\Omega_{t' \rightarrow t}|}
\sum_{p \in \Omega_{t' \rightarrow t}}
\big\|
\hat{D}_{t,p} - \hat{D}_{t' \rightarrow t,p}
\big\|_1
\label{eq:cross_view_loss}
\end{equation}
\vspace{-0.1em}
This encourages depth predictions to be stable under viewpoint changes and respect multi-view constraints, as \cite{pmvc}.

\paragraph{Global Scene-Level Consistency.}
The global regularizer aligns the encoder’s sequence-level representation with that of the frozen 3D foundation model $f_G$~\cite{vggt}. Given the teacher descriptor $f_a$ and the projected encoder descriptor $f_b$ defined above, we use a cosine-distance loss:
\begin{equation}
\mathcal{L}_{\text{global}}
=
1 - \cos(f_a, f_b)
=
1 - \frac{f_a^\top f_b}{\|f_a\|_2 \, \|f_b\|_2}
\label{eq:global_loss}
\end{equation}
\vspace{-0.1em}
Because $f_G$ is frozen and used only at training time, this loss provides lightweight scene-level guidance without introducing any additional inference latency.
\vspace{-0.1em}

\paragraph{Composite Training Objective.}
The total training objective is a sum of the primary $\mathcal{L}_{\text{ce}}$ and our auxiliary losses:


\begin{equation}
\mathcal{L}_{\text{total}}
=
\mathcal{L}_{\text{ce}}
+
\mathcal{L}_{\text{geometry}}
+
\mathcal{L}_{\text{cross-view}}
+
\mathcal{L}_{\text{global}}
\label{eq:total_loss}
\end{equation}
\vspace{-0.1em}
At inference time, the validator $f_P$ and the 3D foundation model $f_G$ are removed, and the MLLM performs 3D-aware reasoning using only the unified encoder features $\{F_t^{3D}\}$ extracted from RGB video.
\vspace{-0.2em}

%% file: tabs/ablation_gt_input.tex
\begin{table}[ht]
\centering
\scriptsize
\setlength{\tabcolsep}{1.5mm}
\renewcommand{\arraystretch}{1.1}
\caption{
Performance gap between 3D-trained models with and without 3D inputs. On 3D VQA benchmarks, removing 3D signals reduces accuracy to the level of zero-shot 2D MLLMs.
}
\resizebox{\linewidth}{!}{

\begin{tabular}{lcccc} 
\toprule
\multirow{2}{*}{Method}
& \multirow{2}{*}{External 3D inputs}
& Scan2Cap 
& ScanRefer
& Multi3DRefer \\

\cmidrule(lr){3-3} \cmidrule(lr){4-4} \cmidrule(lr){5-5}
& & C@0.5 & Acc@0.25  & F1@0.25  \\
\midrule
2D-MLLM (LLaVA-Next-Video)~\cite{llava-next-video} &$\times$ & 31.0 & - & -  \\
3D-MLLM (Video-3D LLM)~\cite{video-llm} &$\times$    & 31.5 & 53.7  & 46.0 \\
3D-MLLM (Video-3D LLM)~\cite{video-llm} &\checkmark  & 83.8 & 58.1  & 58.0  \\


\bottomrule
\end{tabular}
}
\label{tab:remove3D-performance}
\end{table}

%% file: sec/4_experiment.tex
\input{tabs/main_rst}
\section{Experiments}
\label{sec:exp}

In this section, we empirically evaluate 3D-IDE on standard 3D scene understanding benchmarks and analyze how each component of the framework contributes to performance. Throughout, we focus on settings where models receive only RGB inputs at inference, so that any gains stem from the learned 3D-aware representation rather than additional geometric inputs. Unless otherwise specified, our core baseline is a reproduction of Video-3D LLM~\cite{video-llm} with its explicit geometric injection mechanism removed; we denote this baseline as \emph{Video-3D LLM*} in all tables. We first compare 3D-IDE with task-specific specialist models and 3D generalist MLLMs to assess its performance under fair, depth-free conditions and its competitiveness with methods that rely on explicit 3D inputs. We then analyze the learned representation and efficiency of 3D-IDE through geometric correspondence, surface normal estimation, and inference-time measurements. Finally, we conduct ablation studies that examine the impact of geometric supervision, multi-view consistency, and validator initialization, and we provide qualitative visualizations that illustrate the emergent 3D understanding of our model.

\subsection{Datasets and Evaluation Metrics}
\label{sec:datasets_and_metrics}

\begin{figure}[t] 
    \centering
    \begin{subfigure}{\linewidth}
        \centering
        \includegraphics[width=1\linewidth]{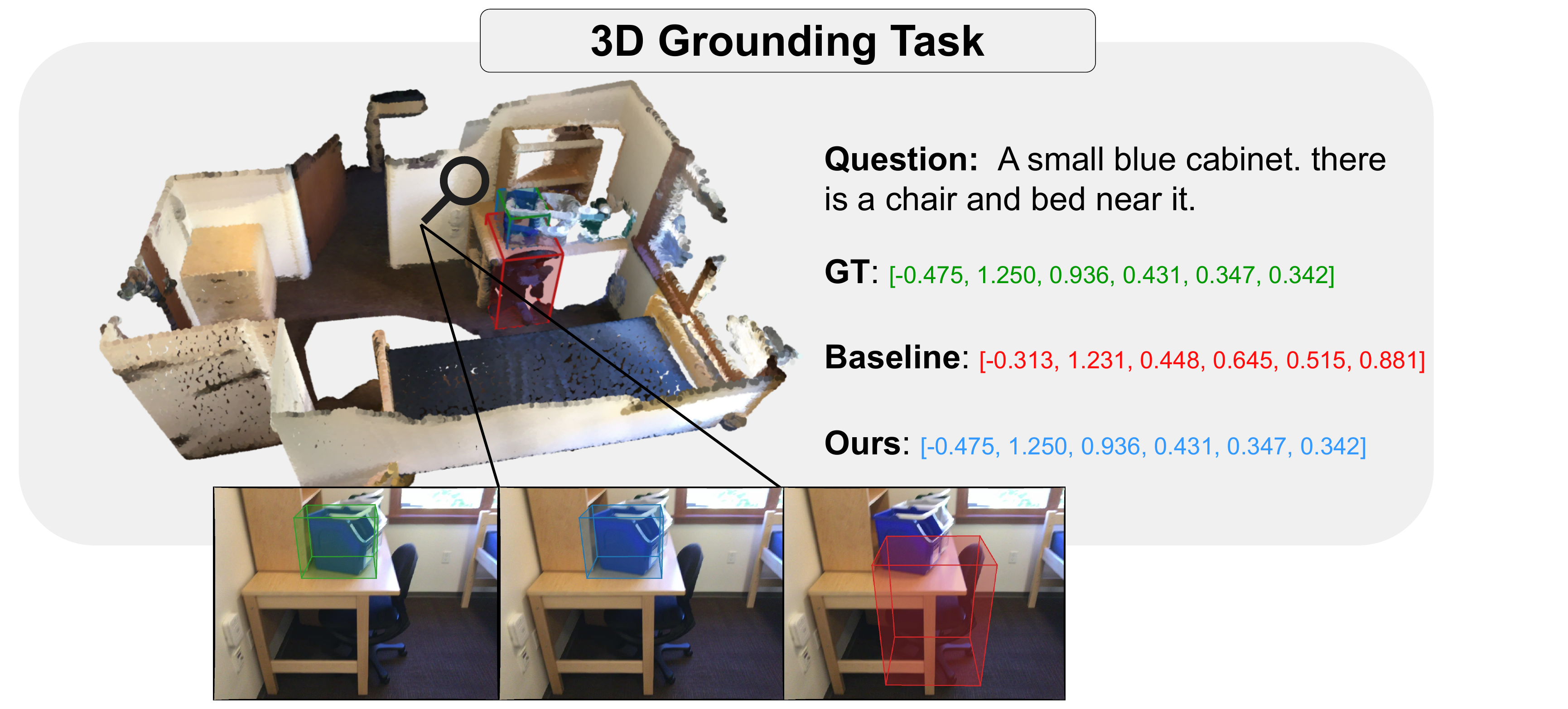}
        \label{fig:grounding}
    \end{subfigure}


    \begin{subfigure}{\linewidth}
        \centering
        \includegraphics[width=1\linewidth]{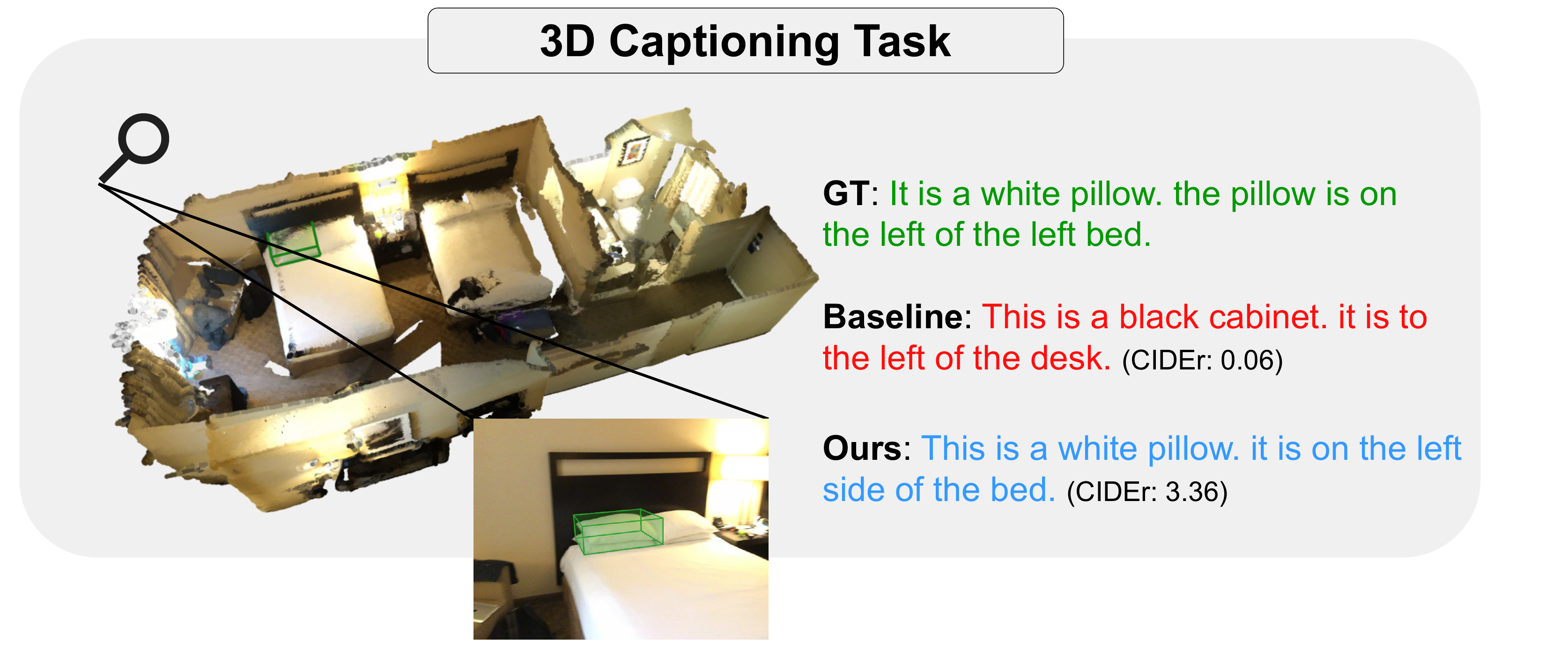}
        \label{fig:captioning}
    \end{subfigure}


    \begin{subfigure}{\linewidth}
        \centering
        \includegraphics[width=1\linewidth]{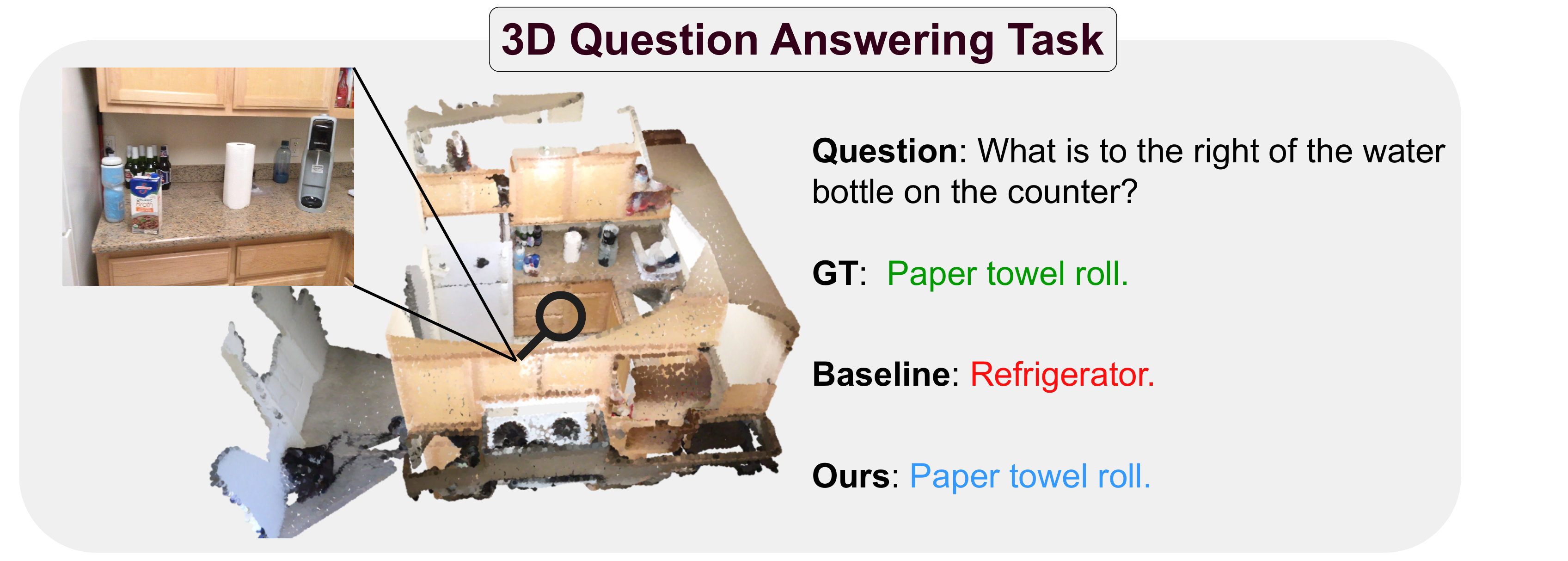}
        \label{fig:vqa}
    \end{subfigure}
    \vspace{-2em}

    \caption{Qualitative results on three 3D vision-language tasks.}
    \label{fig:3d_tasks}
    \vspace{-1em}
\end{figure}

\paragraph{Datasets.}
We evaluate 3D-IDE on five standard 3D vision–language benchmarks, all derived from the ScanNet RGB-D dataset~\cite{scannet}. For 3D visual grounding, we use ScanRefer~\cite{scanrefer} for single-target localization and Multi3DRefer~\cite{multi3drefer} for multi-target localization. For 3D dense captioning, we adopt Scan2Cap~\cite{scan2cap}, which provides region-level descriptions for objects in ScanNet scenes. For 3D question answering, we use ScanQA~\cite{scanqa}, which requires reasoning about spatial relations in indoor environments. To further probe the learned geometric representation, we use NAVI~\cite{navi} for two low-level tasks: geometric correspondence and surface normal estimation.

\noindent\textbf{Evaluation metrics.}
We follow the standard protocols for each benchmark.
For 3D visual grounding, we report Acc@0.25 and Acc@0.5 on ScanRefer, and F1@0.25 and F1@0.5 on Multi3DRefer, all based on 3D IoU thresholds of 0.25 and 0.5.
For 3D dense captioning (Scan2Cap), we report CIDEr and BLEU-4 on predictions whose proposal boxes have IoU $\ge 0.5$ with the ground-truth region, denoted C@0.5 and B-4@0.5.
For 3D question answering, we report Exact Match (EM) accuracy and CIDEr on ScanQA, and EM on SQA3D.
For representation evaluation on NAVI, we follow the Probe3D protocol~\cite{probe3d}: for surface normal estimation we report angular RMSE (lower is better) and mean accuracy (mAcc; higher is better) aggregated over standard angular error thresholds (11.25$^\circ$, 22.5$^\circ$, 30$^\circ$), and for geometric correspondence we report correspondence recall at a 3D point error threshold of 2\,cm (with full recall–threshold curves in the appendix).

\vspace{-0.5em}

\paragraph{Implementation Details.}
Our model, 3D-IDE, is trained end-to-end on 8 NVIDIA H100 GPUs with a batch size of 16. The training process takes around 23 hours for 1 epoch. We use the Adam optimizer~\cite{adam} with an initial learning rate of $2e-6$ for the validator while keeping the learning rates of other modules consistent with the baseline~\cite{video-llm}, using a cosine decay schedule that decays the learning rate down to zero. Following Video-3D LLM, we uniformly sample 32 frames from each scene for both training and evaluation.

\subsection{Main results}

As summarized in \cref{tab:performance}, we compare 3D-IDE with task-specific specialist models that are individually fine-tuned for each benchmark and 3D generalist models that handle diverse multiple tasks within a single framework. Among generalist models, we further distinguish those that require explicit 3D geometric inputs at inference from those that operate purely on RGB inputs. In the RGB-only generalist setting, 3D-IDE achieves the strongest overall results across all three task categories: it sets new state-of-the-art performance on all 3D grounding and QA metrics, and remains highly competitive on Scan2Cap, trailing the best captioning models by less than 3 CIDEr and 1 BLEU-4. At the same time, it is significantly more efficient than representative 3D generalist baselines that incorporate explicit geometry: compared to VG-LLM-8B~\cite{injectvggt}, 3D-IDE reduces the parameter count by $12.86\%$ and inference latency by $55.3\%$ while maintaining comparable accuracy, as reported in \cref{tab:inference-comparison-short}. Moreover, 3D-IDE attains performance comparable to other generalist models that use ground-truth geometric inputs during inference, despite relying only on RGB video. Qualitative results on ScanRefer, Scan2Cap, and ScanQA further illustrate these gains and show that 3D-IDE produces both accurate localizations and descriptions that respect the underlying 3D scene context, as shown in \cref{fig:3d_tasks}

We further evaluate geometric awareness via geometric correspondence and surface normal estimation, as reported in \cref{tab:ablation-representation}. Compared to the RGB-only baseline without IGEP, 3D-IDE achieves higher correspondence recall under both overall and large-baseline settings, and improves both normal RMSE and accuracy. These results indicate that IGEP leads to stronger 3D-aware encoder features, which in turn improve performance on 3D vision–language tasks.


\subsection{Ablation Study}


We conduct an ablation study on ScanRefer and Multi3DRef to quantify the contribution of each IGEP component, as summarized in Tab.~\ref{tab:ablation-table1}. Starting from an RGB-only baseline without auxiliary losses, adding only the global loss $\mathcal{L}_{\text{global}}$ brings modest but consistent gains on both datasets, indicating that aligning the encoder’s sequence-level representation with a frozen 3D foundation model provides useful scene-level regularization. Introducing the geometric loss $\mathcal{L}_{\text{geometry}}$ further improves performance: with $\mathcal{L}_{\text{global}}$ fixed, variants with geometric supervision outperform the baseline by several points on both benchmarks. Comparing the two validator initializations, we observe that the from-scratch validator achieves performance that is very close to, and slightly higher than, its pre-trained counterpart across all reported metrics, showing that IGEP does not rely on a strong pre-trained depth head and that a weak, from-scratch validator is sufficient.

Enabling the cross-view loss $\mathcal{L}_{\text{cross-view}}$ on top of the global and geometric losses yields the best overall configuration, with monotonic improvements across all grounding metrics on both datasets. This demonstrates that localized multi-view consistency complements fine-grained geometric supervision and global sequence-level alignment. Overall, the ablations confirm that each component of IGEP contributes positively and that combining all of them produces the strongest 3D-aware encoder for downstream 3D vision–language tasks. Notably, the full IGEP configuration even surpasses our baseline that relies on Explicit 3D-Input, while using only RGB inputs at inference time.

\input{tabs/ablation2}

\input{tabs/short-geo}

\input{tabs/inference-speed-short}

%% file: tabs/main_rst.tex
\begin{table*}[t]
\centering
\scriptsize
\setlength{\tabcolsep}{1.5mm}
\caption{ Performance comparison on 3D scene understanding benchmarks. Specialists are single-task methods, while generalists are trained for multiple tasks. Bold indicates the best result. Our method belongs to the generalist group without 3D geometric.
}
\resizebox{\linewidth}{!}{
\begin{tabular}{lcccccccccc}
\toprule
\multirow{2}{*}{Method}
& \multicolumn{2}{c}{ScanRefer} 
& \multicolumn{2}{c}{Multi3DRefer} 
& \multicolumn{2}{c}{Scan2Cap}  
& \multicolumn{2}{c}{ScanQA} 
& SQA3D \\ 
\cmidrule(lr){2-3} \cmidrule(lr){4-5} \cmidrule(lr){6-7} \cmidrule(lr){8-9} \cmidrule(lr){10-10}
& Acc@0.25 & Acc@0.5 & F1@0.25 & F1@0.5 & C@0.5 & B-4@0.5 & C & EM & EM \\
\midrule
\multicolumn{10}{l}{\textit{\textbf{Specialists}}} \\
ScanRefer~\cite{scanrefer}     & 37.3 & 24.3 & --   & --   & --   & --   & --   & --   & --   \\
MVT~\cite{mvt}                 & 40.8 & 33.3 & --   & --   & --   & --   & --   & --   & --   \\
3DVG-Trans~\cite{3dvg-trans}   & 45.9 & 34.5 & --   & --   & --   & --   & --   & --   & --   \\
ViL3DRel~\cite{vil3drel}       & 47.9 & 37.7 & --   & --   & --   & --   & --   & --   & --   \\
M3DRef-CLIP~\cite{multi3drefer}& 51.9 & 44.7 & 42.8 & --   & 38.4 & --   & --   & --   & --   \\
Scan2Cap~\cite{scan2cap}       & --   & --   & --   & --   & 35.2 & 22.4 & --   & --   & --   \\
ScanQA~\cite{scanqa}           & --   & --   & --   & --   & --   & --   & 64.9 & 21.1 & 47.2 \\
3D-VisTA~\cite{3dvista}        & 50.6 & 45.8 & --   & --   & 66.9 & 34.0 & 69.6 & 22.4 & 48.5 \\

\midrule
\multicolumn{10}{l}{\textit{\textbf{Generalists} (with 3D geometric inputs)}} \\

3D-LLM(Flamingo)~\cite{3d-llm}           & 21.2 & --   & --   & --   & --   & --   & 59.2 & 20.4 & --   \\
3D-LLM(BLIP2-flant5)~\cite{3d-llm}       & 30.3 & --   & --   & --   & --   & --   & 69.4 & 20.5 & --   \\
Chat-3D~\cite{chat3d}                    & --   & --   & --   & --   & --   & --   & 53.2 & --   & --   \\
Chat-3D v2~\cite{chatscene}              & 42.5 & 38.4 & 45.1 & 41.6 & 63.9 & 31.8 & 87.6 & --   & 54.7 \\
LL3DA~\cite{ll3da}                       & --   & --   & --   & --   & 62.9 & 36.0 & 76.8 & --   & --   \\
SceneLLM~\cite{scenellm}                 & --   & --   & --   & --   & --   & --   & 80.0 & 27.2 & 53.6 \\
LEO~\cite{leo}                           & --   & --   & --   & --   & 72.4 & 38.2 & 101.4& 21.5 & 50.0 \\
Grounded 3D-LLM~\cite{grounded-3dllm}    & 47.9 & 44.1 & 45.2 & 40.6 & 70.6 & 35.5 & 72.7 & --   & --   \\ 
PQ3D~\cite{pq3d}                         & 57.0 & 51.2 & --   & 50.1 & 80.3 & 36.0 & --   & --   & 47.1 \\
ChatScene~\cite{chatscene}               & 55.5 & 50.2 & 57.1 & 52.4 & 77.1 & 36.3 & 87.7 & 21.6 & 54.6 \\
Inst3D-LLM~\cite{inst3d}                 & 57.8 & 51.6 & 58.3 & 53.5 & 79.7 & 38.3 & 88.6 & 24.6 & --   \\
3D-LLaVA~\cite{3dllava}                  & 51.2 & 40.6 & -- & -- & 78.8 & 36.9  & 92.6 & -- & 54.5 \\
LLaVA-3D~\cite{llava3d}                  & 54.1 & 42.4 & --   & --   & 79.2 & 41.1 & 91.7 & 27.0 & 55.6 \\
Video-3D LLM~\cite{video3dllm}           & 58.1 & 51.7 & 58.0 & 52.7 & 83.8 & 41.3 & 102.1 & 30.1  & 58.6 \\
3DRS\cite{3drs}                            & 62.9 & 56.1 & 60.4 & 54.9 & 86.1 & 41.6 & 104.8 & 30.3 & 60.6 \\

\midrule
\multicolumn{10}{l}{\textit{\textbf{Generalists} (without 3D geometric inputs)}} \\
\emph{Video-3D LLM*}~\cite{video3dllm}           & 53.7 & 47.8 & 46.0 & 42.4 & 31.5 &29.9 & 99.7 &29.5  &58.6 \\


VG LLM-4B~\cite{injectvggt}           & 53.5 & 47.5 & -- & -- & 78.6 & 40.9 & -- & --  & 57.0 \\
VG LLM-8B~\cite{injectvggt}           & 57.6 & 50.9 & -- & -- & 80.0 &  \textbf{41.5} & -- & --  & 57.9 \\

VID-LLM~\cite{vid-llm}           & 50.1 & 46.7 & 47.2 & 42.9 & \textbf{81.5} &  40.6 & 101.9 &  27.6  & 57.3 \\

\rowcolor{myblue}
\textbf{Ours}                          & \textbf{60.9} & \textbf{54.5} & \textbf{59.8} & \textbf{54.9} & 79.0 & 40.7 & \textbf{102.1} & \textbf{29.8} & \textbf{59.2} \\
\bottomrule
\end{tabular}
}
\label{tab:performance}
\vspace{-1em}
\end{table*}

%% file: tabs/ablation2.tex
\begin{table}[ht]
\centering
\scriptsize
\setlength{\tabcolsep}{1mm} 
\renewcommand{\arraystretch}{1.1}
\caption{\textbf{Ablation Study.} Effect of the components in our model. $\checkmark$ denotes the component is enabled, $\times$ denotes it is disabled.} 
\begin{tabular}{lll cccc} 
\toprule
\multicolumn{3}{c}{Components} & \multicolumn{2}{c}{ScanRefer} & \multicolumn{2}{c}{Multi3DRef} \\
\cmidrule(lr){1-3} \cmidrule(lr){4-5} \cmidrule(lr){6-7}
Global. & Geometric. & Cross-view. & F1@0.25 & F1@0.5 & Acc@0.25 & Acc@0.5\\
\midrule
$\times$ & $\times$  & $\times$  & 53.7 & 47.8 & 46.0 & 42.4 \\
$\checkmark$ & $\times$  & $\times$ & 56.9 & 50.8 & 55.6 & 51.1 \\
\midrule
$\checkmark$ & $\checkmark$ $\text{pretrain}_\text{vggt}$  & $\times$  & 59.6 & 53.2 & 58.7 & 53.2 \\
$\checkmark$ & $\checkmark$ scratch & $\times$  & 59.8 & 53.3 & 59.7 & 54.3 \\
\midrule

$\checkmark$ & $\checkmark$ $\text{pretrain}_\text{vggt}$ & $\checkmark$  & \textbf{60.5} & \textbf{54.1} & \textbf{59.7} & \textbf{54.5} \\

\rowcolor{myblue}  
$\checkmark$ & $\checkmark$ scratch & $\checkmark$  & \textbf{60.9} & \textbf{54.5} & \textbf{59.8} & \textbf{54.9} \\
\bottomrule

\end{tabular}
\label{tab:ablation-table1}
\end{table}


%% file: tabs/short-geo.tex
\begin{table}[ht]
\centering
\scriptsize
\setlength{\tabcolsep}{3mm} 
\renewcommand{\arraystretch}{1.1}
\caption{\textbf{Geometric Representation Evaluation.}
We compare our method against the baseline on the tasks of Spatial Correspondences and Normal Prediction. Our method demonstrates superior performance by consistently outperforming the baseline across all four metrics. 
Detailed results are available in appendix.
} 

\begin{tabular}{l cccc} 
\toprule
\multicolumn{1}{c}{Method} & \multicolumn{2}{c}{Spatial Correspondences} & \multicolumn{2}{c}{Normal Prediction} \\
\cmidrule(lr){2-3} \cmidrule(lr){4-5} 
 & Recall@2cm $\uparrow$ & $\theta_{90}^{120}\uparrow$ & RMSE $\downarrow$ & mAcc$\uparrow$ \\ 
\midrule
Baseline~\cite{video-llm} & 40.15 & 21.38 & 32.26 & 52.13 \\
\rowcolor{myblue} 
\textbf{Baseline + ours}  & \textbf{42.27} & \textbf{23.06} & \textbf{31.36} & \textbf{53.49} \\
\bottomrule
\end{tabular}
\label{tab:ablation-representation}
\end{table}

%% file: tabs/inference-speed-short.tex
\begin{table}[ht]
\centering
\renewcommand{\arraystretch}{1.1}
\caption{\textbf{Inference Efficiency Comparison.} 
Our model achieves over 2$\times$ faster inference and higher generation throughput while using less GPU memory than VG-LLM under identical settings.}
\resizebox{\linewidth}{!}{%
\begin{tabular}{lcccc}
\toprule
\multicolumn{1}{c}{Method} & Params(B)$\downarrow$ & Mean Time (s)$\downarrow$ & Tokens/s $\uparrow$ & Peak Mem(GB)$\downarrow$ \\
\midrule
VG LLM-8B~\cite{injectvggt} & 9.25 & 3.60 & 4.32 & 21.10 \\
\rowcolor{myblue}
\textbf{Ours} & \textbf{8.06} & \textbf{1.61} & \textbf{10.72} & \textbf{18.35} \\
\bottomrule
\end{tabular}%
}
\label{tab:inference-comparison-short}
\end{table}

%% file: sec/5_conclusion.tex
\vspace{-0.5em}
\section{Conclusion}
\label{sec:conclusion}
We presented 3D-IDE, a framework that addresses the inference-time latency and data dependencies of 3D-aware MLLMs. Guided by the IGEP, a lightweight training-only validator encourages a unified RGB encoder to internalize 3D structure from video. As a result, 3D-IDE matches or surpasses state-of-the-art performance while requiring no 3D inputs or auxiliary encoders at inference and incurring no latency overhead. This shows that 3D knowledge can be learned implicitly within a single encoder, enabling more practical 3D understanding models.

%% file: sec/X_suppl.tex
\clearpage
\setcounter{page}{1}
\maketitlesupplementary

\section{Datasets Statistics}
\noindent \textbf{Training data.}
For fine-tuning, we adopt the same pool of 3D Understanding and Reasoning benchmarks as Video-3D LLM~\cite{video3dllm}, namely ScanRefer, Multi3DRefer, Scan2Cap, ScanQA, and SQA3D.
In total, this yields $223{,}128$ training examples: SQA3D provides $79{,}445$ samples ($35.6$\% of the corpus), Multi3DRefer $43{,}838$ ($19.6\%$), ScanRefer and Scan2Cap $36{,}665$ each ($16.4\%$ per dataset), and ScanQA $26{,}515$ ($11.9\%$).
All datasets except SQA3D are built on $562$ reconstructed scans, while SQA3D covers $518$ scans.
The average question length ranges from 13 to 38 words across datasets.
Scan2Cap and ScanQA additionally offer answer sentences averaging $17.9$ and $2.4$ words, and SQA3D has relatively long questions ($37.8$ words on average) with very short answers ($1.1$ words).

\vspace{5pt}
\noindent \textbf{Evaluation data.}
For evaluation, we use the official validation splits of ScanRefer, Multi3DRefer, Scan2Cap, and ScanQA, together with the test split of SQA3D. The combined evaluation suite contains $30{,}890$ instances: $11{,}120$ from Multi3DRefer ($36.0\%$), $9{,}508$ from ScanRefer ($30.8\%$), $4{,}675$ from ScanQA ($15.1\%$), $3{,}519$ from SQA3D ($11.4\%$), and $2{,}068$ from Scan2Cap ($6.7\%$).
The average question length in these splits varies between $13.0$ and $36.3$ words, while Scan2Cap and ScanQA provide answer texts with mean lengths of $18.7$ and $2.4$ words, respectively; SQA3D again features long questions ($36.3$ words) with very short answers ($1.1$ words).

\section{Additional Ablative Analysis}

\noindent \textbf{Geometric Representation.}
As shown in \cref{tab:corr_navi,tab:snorm_navi}, both the pretrained-head and from-scratch-head variants improve encoder probing scores compared to training without any validator, confirming that geometric supervision is beneficial.
Among them, the from-scratch validator yields the highest normal and correspondence accuracy, indicating that it encourages the encoder to internalize 3D structure more effectively than relying on a stronger pretrained head.
This outcome is consistent with the design philosophy of IGEP: the geometric validator is intentionally kept weak and low-capacity so that it cannot absorb complex 3D reasoning on its own. To minimize the geometric loss, the visual encoder is instead pressured to internalize 3D structure within the shared tokens so that this low-capacity validator can decode it. Geometry therefore neither explicitly injected nor disentangled, but emerges under optimization pressure within a unified representation space.
For surface normal estimation, we follow the Probe3D~\cite{probe3d} protocol and train linear probing heads on features extracted from the frozen visual encoder. The reported geometry is thus derived entirely from the implicit RGB tokens; the training-only validator is not involved at evaluation time.

\begin{figure}[t]
    \centering
    \begin{subfigure}{\linewidth}
        \centering
        \includegraphics[width=1.0\linewidth]{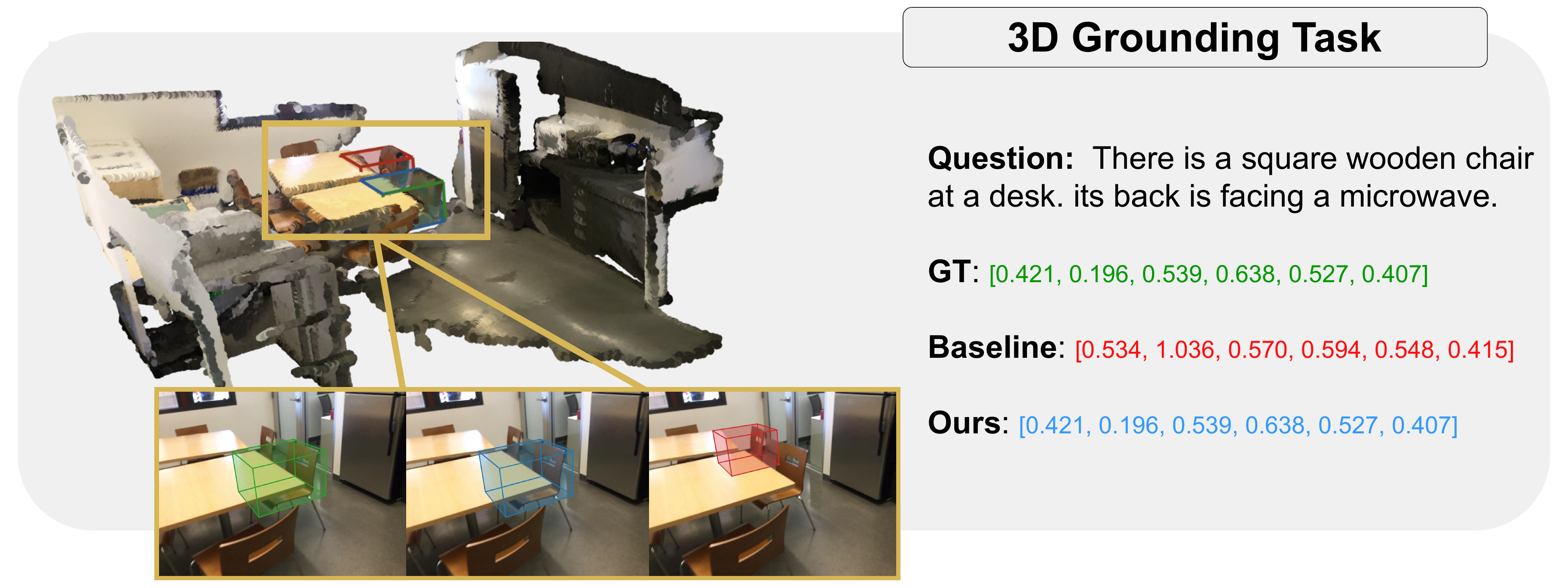}
    \end{subfigure}

    \vspace{0.8em}

    \begin{subfigure}{\linewidth}
        \centering
        \includegraphics[width=1.0\linewidth]{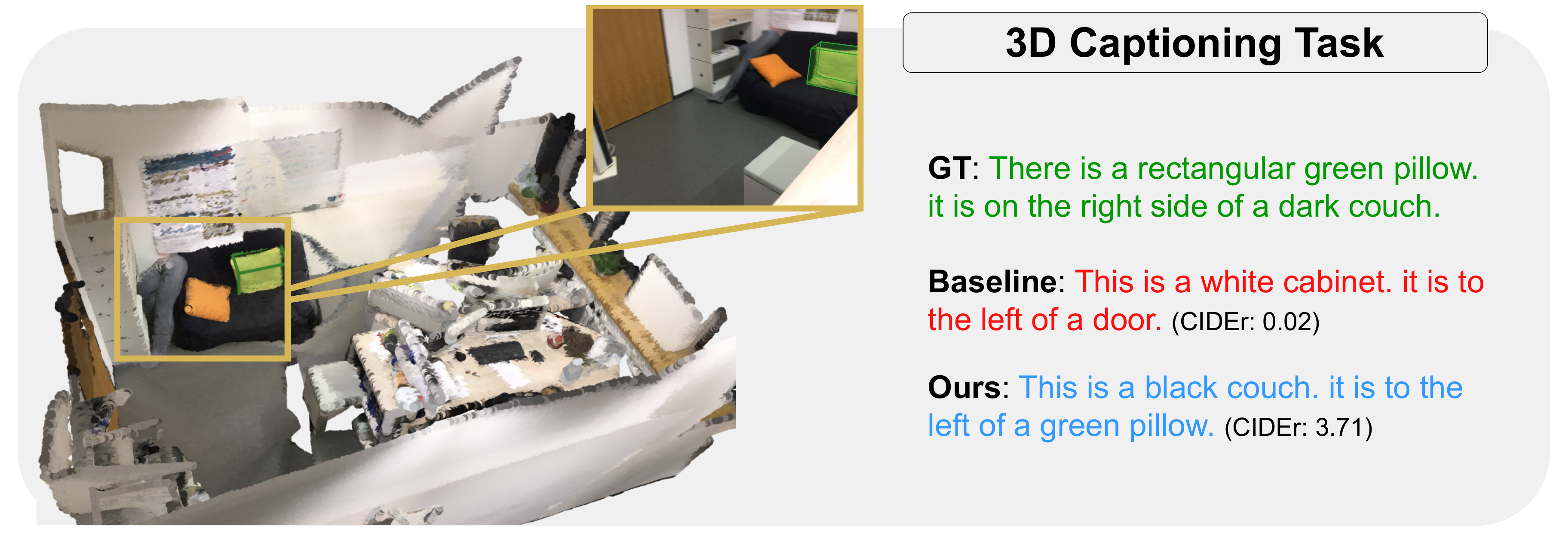}
    \end{subfigure}

    \vspace{0.8em}

    \begin{subfigure}{\linewidth}
        \centering
        \includegraphics[width=1.0\linewidth]{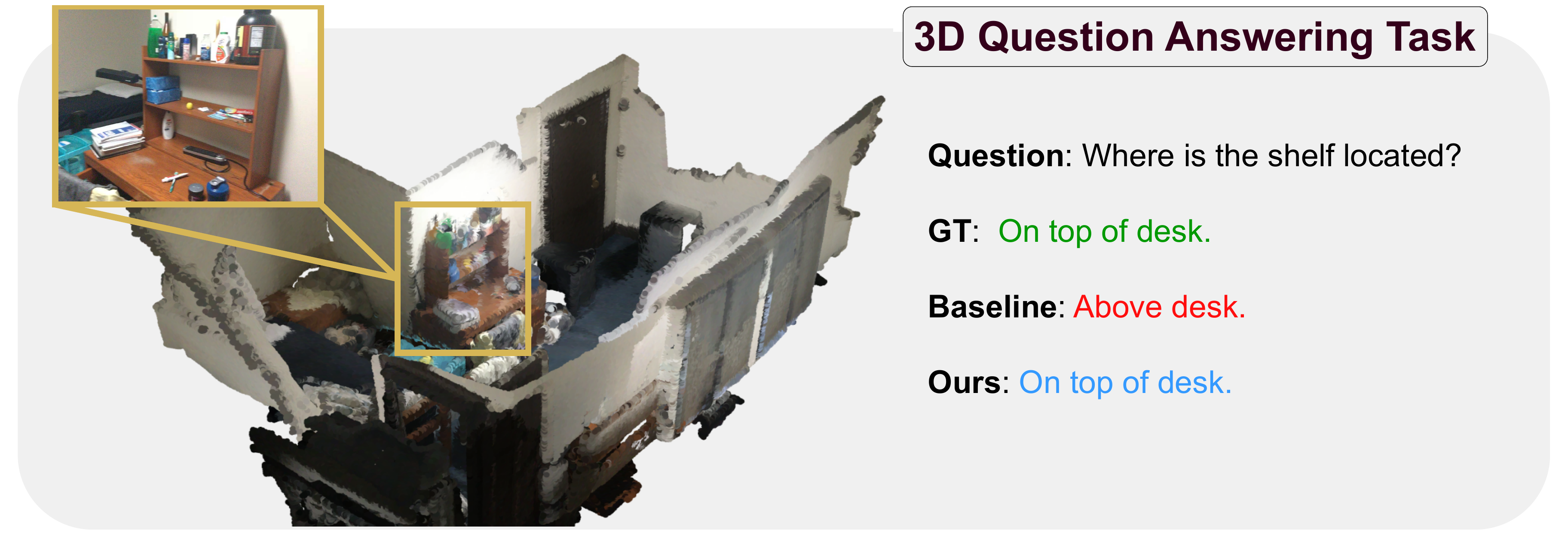}
    \end{subfigure}

    \caption{More qualitative results on three 3D vision-language tasks: language-guided object localization (top), region-level captioning (middle), and spatial question answering (bottom). In the grounding examples, green 3D bounding boxes denote the ground-truth targets, red boxes the predictions of the baseline, and blue boxes the predictions of our model. Our method better aligns with the targets and produces more accurate captions and answers.}
    \label{fig:3d_tasks_appendix1}
    \vspace{-10pt}
\end{figure}

\input{tabs/ablation_global}

\begin{figure}[t]
    \centering
    \includegraphics[width=\linewidth]{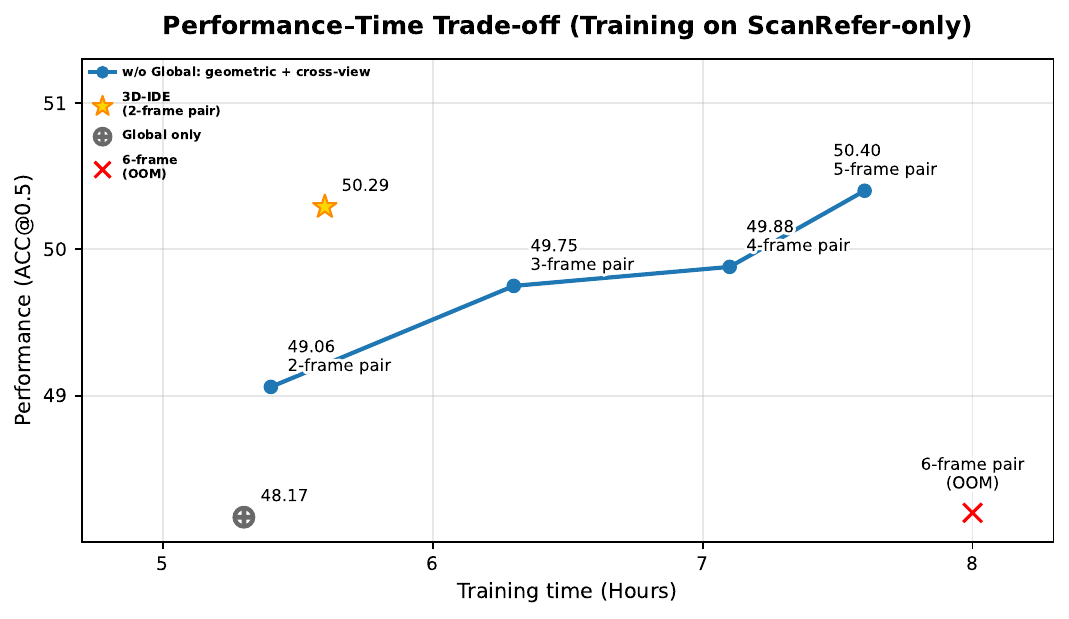}
    \vspace{-1em}
    \caption{3D-IDE retains performance w/o global supervision, at the cost of longer training and higher VRAM (OOM at 6-frame).}
    \label{fig:trade-off}
    \vspace{-1.5em}
\end{figure}

\vspace{5pt}
\noindent \textbf{Role of Global Supervision.}
A natural concern is that 3D-IDE might simply inherit 3D knowledge from the foundation model $f_G$~\cite{vggt} through the global supervision. To disentangle this effect, we ablate $\mathcal{L}_{\text{global}}$ in \cref{tab:ablation-global}. Using only local geometric and cross-view constraints, \ie, $\mathcal{L}_{\text{geometry}}$ and $\mathcal{L}_{\text{cross-view}}$ with 2-frame warping, already yields a substantial gain over the baseline and even surpasses using $\mathcal{L}_{\text{global}}$ alone. This indicates that the principal source of 3D awareness arises from IGEP itself rather than being dictated by $f_G$. At the same time, combining local and global supervision achieves the best overall performance, while $\mathcal{L}_{\text{global}}$ is used at training and is entirely discarded at inference. In practice, the global term behaves as a training-time scene-level regularizer that approximates dense multi-view constraints whose pairwise complexity grows quadratically with sequence length, delivering an almost ``free-lunch'' improvement in 3D consistency without any inference latency.
Importantly, $f_G$ and the geometric validator operate on distinct parts of the architecture: $f_G$ supervises only the final VLM hidden space via $\mathcal{L}_{\text{global}}$, whereas the geometric validator acts solely on the upstream shared encoder via $\mathcal{L}_{\text{geometry}}$ and $\mathcal{L}_{\text{cross-view}}$. This separation avoids direct coupling between the teacher and validator feature spaces, reducing the risk of misaligned supervision signals.
As shown in \cref{fig:trade-off}, 3D-IDE retains strong performance even without global supervision, though at the cost of longer training and higher VRAM usage.

\section{Detailed Comparison}
\vspace{-0.5em}
Here, we conduct a thorough comparison with other methods, covering all metrics across five benchmark tasks.

\begin{table}[h]
\centering
\scriptsize
\setlength{\tabcolsep}{4pt}
\renewcommand{\arraystretch}{1.1}
\begin{tabular}{lccccc}
\toprule
Components
  & ScanRefer            & Multi3DRefer         & Scan2Cap             & ScanQA               & SQA3D \\
  & Acc@0.25\,$\uparrow$ & F1@0.25\,$\uparrow$  & CIDEr\,$\uparrow$    & CIDEr\,$\uparrow$    & EM\,$\uparrow$ \\
\midrule
Baseline                                      & 53.7 & 46.0 & 31.5  & 99.7  & 58.6 \\
$+\,\mathcal{L}_{\text{global}}$              & 56.9 & 55.6 & 77.7  & 100.0 & 57.8 \\
$+\,\mathcal{L}_{\text{geometry}}$            & 59.8 & 59.7 & 78.7  & 101.9 & 59.0 \\
\rowcolor{myblue}
$+\,\mathcal{L}_{\text{cross-view}}$ (Full)  & \textbf{60.9} & \textbf{59.8} & \textbf{79.0} & \textbf{102.1} & \textbf{59.2} \\
\bottomrule
\end{tabular}
\vspace{0.5em}
\caption{Extended ablation across all five benchmarks. Each row cumulatively adds one IGEP component. The full model consistently achieves the best results across all tasks.}
\label{tab:ablation5}
\end{table}

\vspace{5pt}
\noindent \textbf{Extended Ablation Across All Benchmarks.}
To further validate the contribution of each component, \cref{tab:ablation5} extends the ablation in the main paper to all five benchmarks by cumulatively adding each objective. Each component brings consistent improvements across tasks, and the full configuration achieves the best results on all five benchmarks.

\vspace{4pt}
\noindent \textbf{Impact of Removing 3D Inputs from 3DRS.}
A key claim of our work is that methods relying on ground-truth depth and camera pose at inference time suffer a severe performance drop when those inputs are withheld. \cref{tab:rgbonly} substantiates this by comparing 3DRS~\cite{3drs} in its original setting (with 3D geometric inputs) against its RGB-only variant (3DRS*, without 3D inputs) and our method, which is RGB-only by design. Removing 3D inputs causes a sharp drop in 3DRS performance on both ScanRefer and Multi3DRefer, whereas 3D-IDE remains strong and outperforms 3DRS* by a substantial margin on all grounding metrics. This confirms that 3DRS cannot be categorized as a generalist model without 3D geometric inputs, as it requires ground-truth depth and camera pose at inference to construct coordinate maps.

\begin{table}[h]
\centering
\scriptsize
\setlength{\tabcolsep}{5pt}
\renewcommand{\arraystretch}{1.1}
\begin{tabular}{lc cccc}
\toprule
\multirow{2}{*}{Method} & \multirow{2}{*}{3D inputs} &
  \multicolumn{2}{c}{ScanRefer} & \multicolumn{2}{c}{Multi3DRefer} \\
\cmidrule(lr){3-4}\cmidrule(lr){5-6}
  & & Acc@0.25 & Acc@0.5 & F1@0.25 & F1@0.5 \\
\midrule
3DRS~\cite{3drs}  & \checkmark & 62.9  & 56.1  & 60.4 & 54.9 \\
3DRS*~\cite{3drs} & $\times$   & 56.95 & 50.83 & 55.8 & 51.1 \\
\rowcolor{myblue}
\textbf{3D-IDE (Ours)}             & $\times$   & \textbf{60.9} & \textbf{54.5} & \textbf{59.8} & \textbf{54.9} \\
\bottomrule
\end{tabular}
\vspace{0.5em}
\caption{Effect of removing 3D geometric inputs from 3DRS at inference. 3DRS* denotes the RGB-only variant. Our method closes most of the gap to the full 3DRS while using no 3D inputs.}
\label{tab:rgbonly}
\end{table}

\vspace{4pt}
\noindent \textbf{ScanRefer.}
As shown in the detailed ScanRefer~\cite{scanrefer} results in \cref{tab:scanrefer_detail}, our method achieves strong overall performance, with clear improvements over the baseline on both Acc@0.25 and Acc@0.5, indicating better fine-grained localization of the target object.

\vspace{4pt}
\noindent \textbf{Multi3DRefer.}
Following \cite{multi3drefer}, we evaluate all question types, including zero-target (ZT), single-target (ST), and multi-target (MT) cases, with and without distractors. From \cref{tab:multi3drefer_detail}, our approach consistently outperforms previous methods on the ST and MT splits under both distractor settings, demonstrating stronger robustness to spurious objects. Interestingly, the depth-free variants of Video 3D-LLM and 3DRS obtain higher ZT scores but substantially worse ST and MT results, indicating a bias toward predicting no target once geometric cues are removed.

\vspace{4pt}
\noindent \textbf{ScanQA.}
On the ScanQA validation set~\cite{scanqa}, our method achieves better results than prior approaches on key metrics such as EM@1 and CIDEr, and is competitive on BLEU and METEOR, as shown in \cref{tab:scanqa_detail}. These results highlight the effectiveness of our model for 3D question answering.

\vspace{4pt}
\noindent \textbf{SQA3D.}
As shown in \cref{tab:sqa3d_detail}, our method establishes new state-of-the-art performance on the SQA3D test split~\cite{sqa3d}. It attains the highest overall EM and consistently improves over previous methods across most question types, indicating strong generalization to diverse question categories.

\vspace{4pt}
\noindent \textbf{Scan2Cap.}
On the Scan2Cap validation benchmark~\cite{scan2cap}, we adopt the training and inference protocol of~\cite{video-llm}.
Under this setting, our method substantially improves over our baseline and attains CIDEr and BLEU-4 scores close to the best results, while remaining competitive on METEOR and ROUGE-L, as summarized in \cref{tab:scan2cap_detail}.

\vspace{-0.3em}
\begin{figure}[t]
    \centering
    \begin{subfigure}{\linewidth}
        \centering
        \includegraphics[width=1\linewidth]{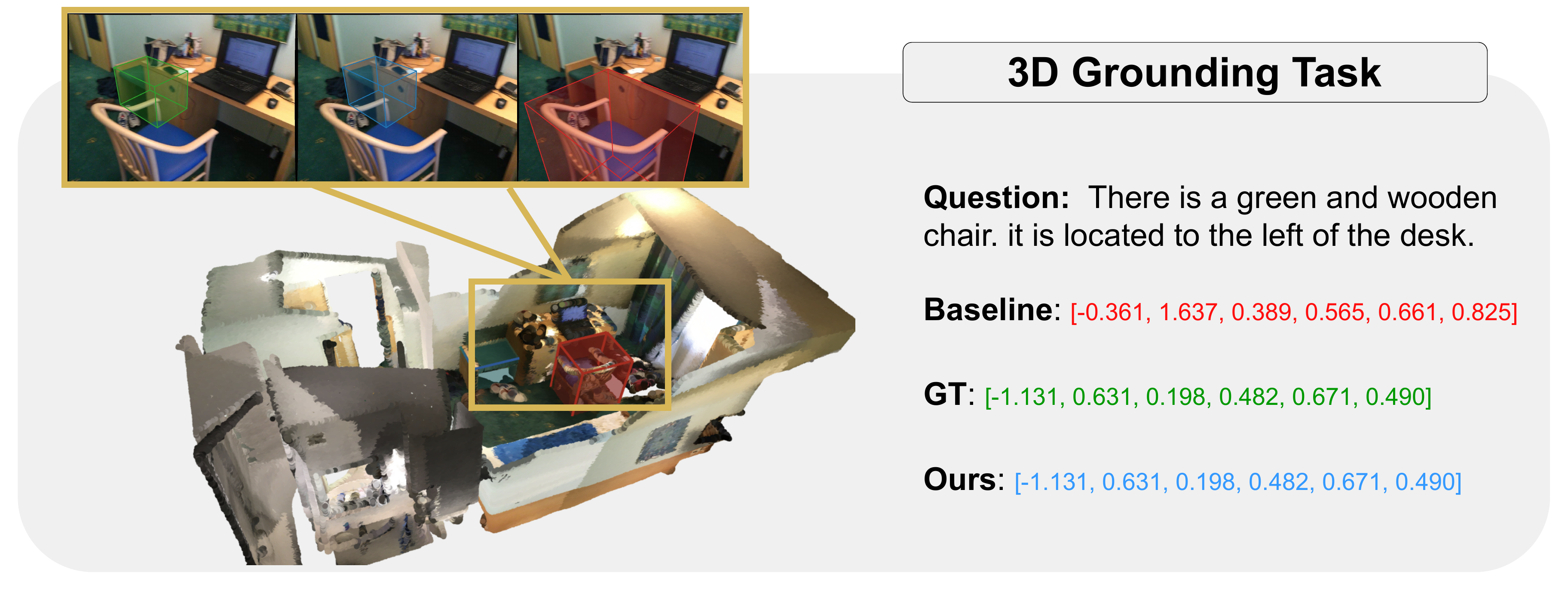}
    \end{subfigure}

    \vspace{0.6em}

    \begin{subfigure}{\linewidth}
        \centering
        \includegraphics[width=1\linewidth]{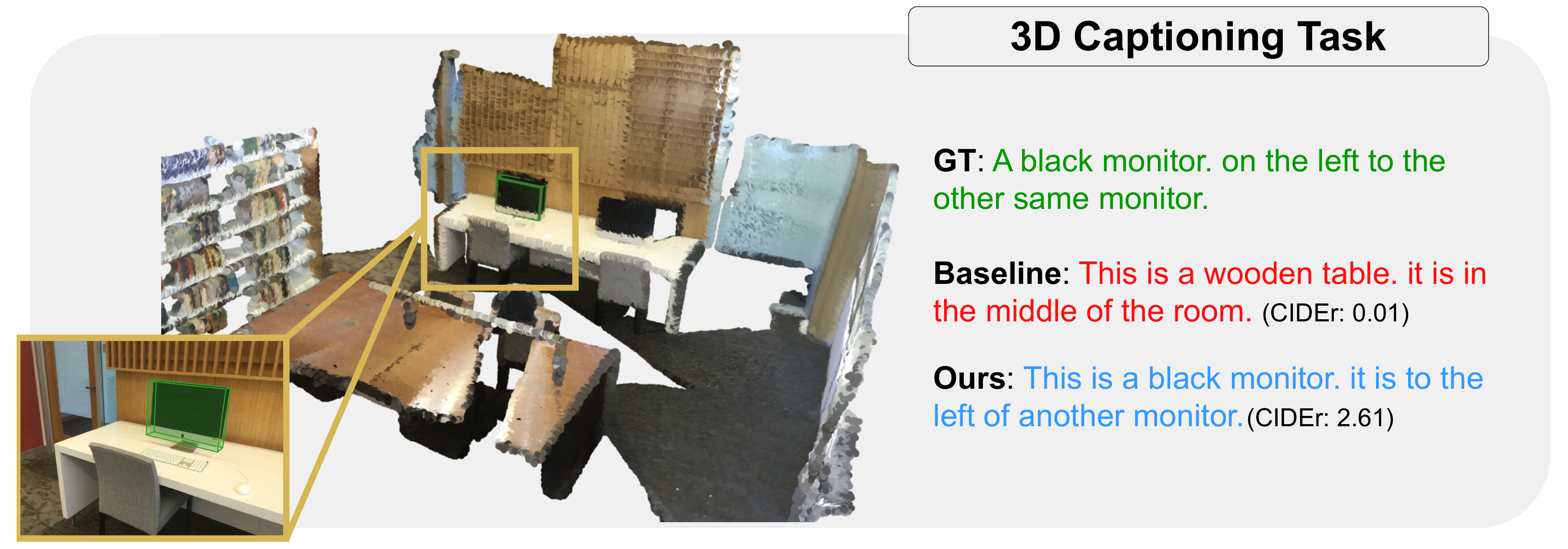}
    \end{subfigure}

    \vspace{0.6em}

    \begin{subfigure}{\linewidth}
        \centering
        \includegraphics[width=1\linewidth]{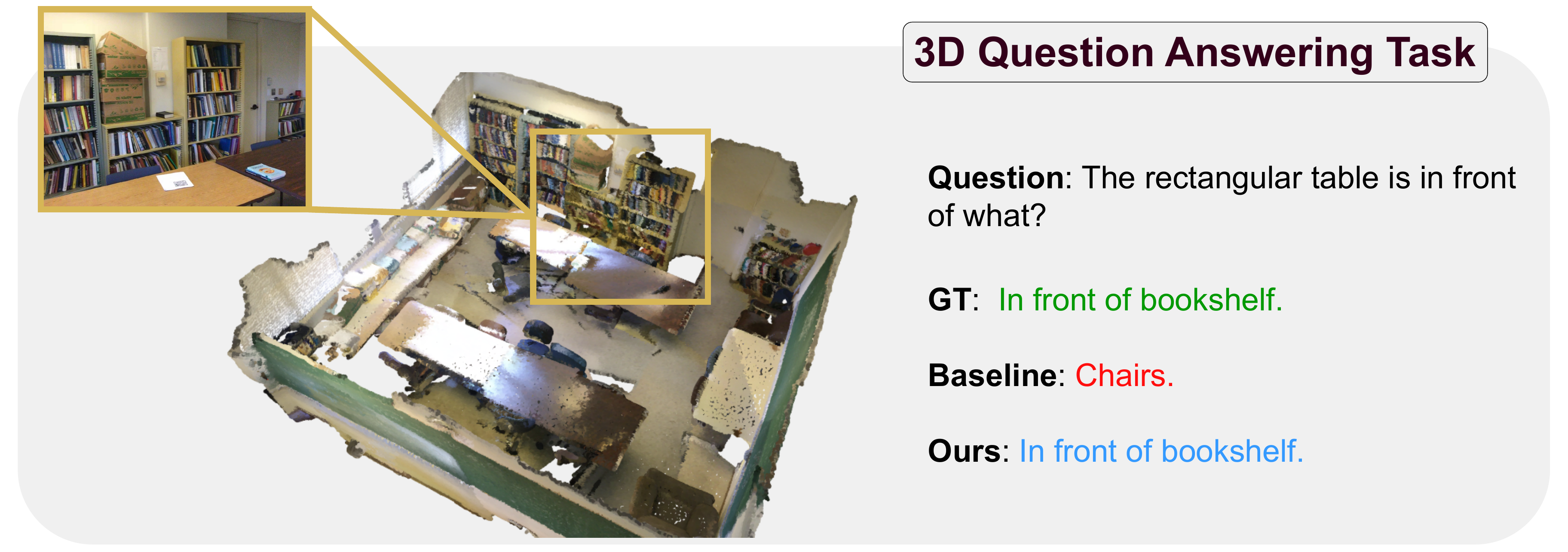}
    \end{subfigure}

    \caption{More qualitative results on three 3D vision-language tasks (continued). The three rows correspond to object localization, region-level captioning, and spatial question answering, respectively. Color coding is the same as in Figure~\ref{fig:3d_tasks_appendix1}. Our model remains effective across diverse scenes and linguistic queries.}
    \label{fig:3d_tasks_appendix2}
\end{figure}

\section{More Qualitative Results}
\vspace{-0.5em}
\cref{fig:3d_tasks_appendix1,fig:3d_tasks_appendix2} qualitatively summarize the behavior of our model on three challenging 3D scene understanding tasks: language-guided object localization, region-level captioning, and spatial question answering.
In the visual grounding examples, the model must retrieve the correct object in a cluttered 3D environment given a natural-language description. For each case we visualize three bounding boxes: green denotes the ground-truth target, red the prediction of the RGB-only baseline, and blue the prediction of our model. Our predictions align much more closely with the intended targets, indicating that the model can reliably interpret both spatial and semantic cues from language.
Across all three tasks, these qualitative results demonstrate that, even without any 3D input during inference, our method leverages its learned 3D-aware representation to produce more accurate and coherent outputs than the baseline.

\input{tabs/scan2cap_detail}
\input{tabs/sqa3d_detail}

\input{tabs/rep_2_correspon}

\input{tabs/rep_2_normal}

\input{tabs/scanrefer_detail}
\input{tabs/multirefer_detail}

\vspace{-10em}
\input{tabs/scanqa_detail}

\clearpage

%% file: tabs/ablation_global.tex
\begin{table}[b]
\centering
\scriptsize
\setlength{\tabcolsep}{1mm} 
\renewcommand{\arraystretch}{1.1}
\caption{\textbf{Additional Ablation Study.} Effect of $\mathcal{L}_{\text{global}}$ to speed-up training. $\checkmark$ denotes enabled, $\times$ denotes disabled.} 
\begin{tabular}{lll cccc} 
\toprule
\multicolumn{3}{c}{Components} & \multicolumn{2}{c}{ScanRefer} & \multicolumn{2}{c}{Multi3DRef} \\
\cmidrule(lr){1-3} \cmidrule(lr){4-5} \cmidrule(lr){6-7}
Global. & Geometric. & Cross-view. & Acc@0.25 & Acc@0.5 & F1@0.25 & F1@0.5\\
\midrule

$\times$ & $\times$  & $\times$  & 53.7 & 47.8 & 46.0 & 42.4 \\
\midrule

$\checkmark$ & $\times$  & $\times$ & 56.9 & 50.8 & 55.6 & 51.1 \\
$\times$ & $\checkmark$  & $\checkmark$ & 58.6 & 51.9 & 57.8 & 52.5 \\

\midrule
$\checkmark$ & $\checkmark$ scratch & $\times$  & 59.8 & 53.3 & 59.7 & 54.3 \\

\rowcolor{myblue}  
$\checkmark$ & $\checkmark$ scratch & $\checkmark$  & \textbf{60.9} & \textbf{54.5} & \textbf{59.8} & \textbf{54.9} \\
\bottomrule

\end{tabular}
\label{tab:ablation-global}
\end{table}

%% file: tabs/scan2cap_detail.tex
\begin{table}[t]
\centering
\caption{Performance comparison on the \textbf{Scan2Cap} validation set.}
\label{tab:scan2cap_detail}
\resizebox{\linewidth}{!}{%
\begin{tabular}{lcccc}
\toprule
\multirow{2}{*}{Method} & \multicolumn{4}{c}{@0.5} \\
 & C & B-4 & M & R \\
\midrule
Scan2Cap~\cite{scan2cap} & 39.08 & 23.32 & 21.97 & 44.48 \\
3D-VisTA~\cite{3dvista} & 66.90 & 34.00 & 27.10 & 54.30 \\
ChatScene~\cite{chatscene} & 77.19 & 36.34 & 28.01 & 58.12 \\
LLaVA-3D~\cite{llava3d} & 79.21 & 41.12 & 30.21 & 63.41 \\
\midrule
baseline~\cite{video3dllm} & 31.53 & 29.98 & 24.18 & 57.66 \\
VG-LLM~\cite{injectvggt} & 80.00 & 41.50 & 28.90 & 62.60 \\
\rowcolor{myblue}
\textbf{Ours} & 79.02 & 40.76 & 28.79 & 62.13 \\
\bottomrule
\end{tabular}%
}
\end{table}

%% file: tabs/sqa3d_detail.tex


\begin{table}[t]
\centering
\small
{\setlength{\tabcolsep}{1mm}%
\begin{tabular*}{\linewidth}{@{\extracolsep{\fill}}lccccccc}
\toprule
\multirow{2}{*}{Method} & \multicolumn{6}{c}{Test set} & \multirow{2}{*}{\textbf{Avg.}} \\ \cmidrule{2-7}
 & What & Is & How & Can & Which & Others &  \\
\midrule
3D-VisTA~\cite{3dvista}   & 34.8 & 63.3 & 45.4 & 69.8 & 47.2 & 48.1 & 48.5 \\
Scene-LLM~\cite{scenellm} & 40.9 & 69.1 & 45.0 & 70.8 & 47.2 & 52.3 & 54.2 \\
ChatScene~\cite{chatscene}& 45.4 & 67.0 & 52.0 & 69.5 & 49.9 & 55.0 & 54.6 \\
LLaVA-3D~\cite{llava3d}   & --   & --   & --   & --   & --   & --   & 55.6 \\

Video-3D~\cite{video3dllm} & 51.1 & 72.4 & 55.5 & 69.8 & {51.3} & 56.0 & 58.6 \\
\midrule

baseline~\cite{video3dllm} & 51.8   & 73.1   & 56.5   & 70.1   &  51.0   &54.7   & 58.5  \\


\rowcolor{myblue}
\textbf{Ours} &  \textbf{51.8}   & {72.7}   & \textbf{60.4}   & {68.3}   & {49.0}   & \textbf{58.0}   &  \textbf{59.2}\\

\bottomrule
\end{tabular*}}
\caption{Performance comparison on the test set of \textbf{SQA3D}.}
\label{tab:sqa3d_detail}
\vspace{-2em}
\end{table}

%% file: tabs/rep_2_correspon.tex
\begin{table*}
  \centering
  \renewcommand{\arraystretch}{1.0}
  \caption{\textbf{Correspondence Estimation Results for NAVI.} We present the NAVI correspondence estimation results for all models. The results are presented for features extracted at different layers with performance binned for different relative viewpoint changes between image pairs. The highest performing entry in each column is bolded.}
  \label{tab:corr_navi}
  \resizebox{\linewidth}{!}{%
  \begin{tabular}{
    l l
    @{\hskip 10pt} r rrr
    @{\hskip 10pt} r rrr
    @{\hskip 10pt} r rrr
    @{\hskip 10pt} r rrr
  }
    \toprule
    & &
    \multicolumn{4}{c}{Block$_0$} &
    \multicolumn{4}{c}{Block$_1$} &
    \multicolumn{4}{c}{Block$_2$} &
    \multicolumn{4}{c}{Block$_3$} \\
    \cmidrule(lr){3-6}
    \cmidrule(lr){7-10}
    \cmidrule(lr){11-14}
    \cmidrule(lr){15-18}
    \textbf{Model} & \textbf{Venue} &
    $\theta_{0}^{30}$ & $\theta_{30}^{60}$ & $\theta_{60}^{90}$ & $\theta_{90}^{120}$ &
    $\theta_{0}^{30}$ & $\theta_{30}^{60}$ & $\theta_{60}^{90}$ & $\theta_{90}^{120}$ &
    $\theta_{0}^{30}$ & $\theta_{30}^{60}$ & $\theta_{60}^{90}$ & $\theta_{90}^{120}$ &
    $\theta_{0}^{30}$ & $\theta_{30}^{60}$ & $\theta_{60}^{90}$ & $\theta_{90}^{120}$ \\
    \midrule
    Video-3D LLM~\cite{video-llm} & CVPR'25 &
    \textbf{75.99} & \textbf{38.83} & \textbf{20.27} & 10.68 &
    80.48 & 52.97 & 32.15 & 17.70 &
    75.36 & 49.26 & 34.94 & 21.38 &
    71.97 & 46.28 & 34.50 & 22.19 \\
    3DRS~\cite{3drs} & NIPS'25 &
    74.05 & 37.67 & 19.81 & 10.60 &
    79.59 & 52.13 & \textbf{33.07} & \textbf{17.87} &
    73.88 & 48.77 & 35.76 & 21.89 &
    69.64 & 45.57 & 34.86 & 22.69 \\
    \textbf{3D-IDE (Ours)}$_\text{pretrain}$ & {--} &
    74.63 & 38.07 & 19.99 & 10.69 &
    81.60 & 53.64 & 33.03 & 17.33 &
    77.04 & \textbf{52.25} & 37.13 & 22.34 &
    72.33 & 47.68 & 35.15 & 22.11 \\
    \rowcolor{myblue}
    \textbf{3D-IDE (Ours)}$_\text{scratch}$ & {--} &
    74.93 & 38.05 & 19.89 & \textbf{10.77} &
    \textbf{81.63} & \textbf{53.79} & 33.06 & 17.38 &
    \textbf{77.05} & 52.02 & \textbf{37.58} & \textbf{23.06} &
    \textbf{72.46} & \textbf{47.97} & \textbf{36.16} & \textbf{23.21} \\
    \bottomrule
  \end{tabular}%
  }
\end{table*}

%% file: tabs/rep_2_normal.tex
\sisetup{detect-weight=true,detect-inline-weight=math,table-number-alignment=center}
\newcolumntype{A}{S[table-format=2.2]}
\newcolumntype{R}{S[table-format=2.2]}
\begin{table*}[!ht]
  \centering
  \renewcommand{\arraystretch}{1.12}
  \setlength{\aboverulesep}{0.3ex}
  \setlength{\belowrulesep}{0.3ex}
  \caption{\textbf{Surface Normal Estimation Results.} We present the surface normal estimation results for all models. Higher is better for accuracy, lower is better for RMSE. The highest performing entry in each column is bolded.}
  \label{tab:snorm_navi}
  \resizebox{\linewidth}{!}{%
  \begin{tabular}{l l @{\hskip 10pt} A A A R R}
    \toprule
    & & \multicolumn{5}{c}{\textbf{NAVI (Test)}} \\
    \cmidrule(lr){3-7}
    \textbf{Model} & \textbf{Venue} &
      \multicolumn{1}{c}{Acc@11.25$^\circ$ (\%)} &
      \multicolumn{1}{c}{Acc@22.5$^\circ$ (\%)} &
      \multicolumn{1}{c}{Acc@30$^\circ$ (\%)} &
      \multicolumn{1}{c}{RMSE ($^\circ$) $\downarrow$} &
      \multicolumn{1}{c}{mAcc (\%) $\uparrow$} \\
    \midrule
    Video-3D LLM~\cite{video-llm} & CVPR'25 &
      28.69 & 57.65 & 70.06 & 32.26 & 52.13 \\
    3DRS~\cite{3drs} & NIPS'25 &
      28.61 & 57.90 & 70.48 & 31.86 & 52.33 \\
    \textbf{3D-IDE (Ours)}$_\text{pretrain}$ & {--} &
      29.85 & 58.73 & 71.05 & 31.46 & 53.21 \\
    \rowcolor{myblue}
    \textbf{3D-IDE (Ours)}$_\text{scratch}$ & {--} &
      \bfseries 30.15 & \bfseries 59.01 & \bfseries 71.32 & \bfseries 31.46 & \bfseries 53.49 \\
    \bottomrule
  \end{tabular}%
  }
\end{table*}

%% file: tabs/scanrefer_detail.tex
\begin{table*}[b]
    \centering
    \caption{\textbf{Performance comparison on the validation set of ScanRefer.}
    ``Unique'' and ``Multiple'' depends on whether there are other objects of the same class as the target object.
211    } 
    {\small
    \resizebox{\textwidth}{!}{
    \begin{tabular}{lcccccc}
        \toprule
        \multirow{2}{*}{Method} & \multicolumn{2}{c}{Unique} & \multicolumn{2}{c}{Multiple} & \multicolumn{2}{c}{Overall} \\
         & Acc@0.25 & Acc@0.5 & Acc@0.25 & Acc@0.5 & Acc@0.25 & Acc@0.5 \\
         \midrule
        ScanRefer~\cite{scanrefer} & 76.3 & 53.5 & 32.7 & 21.1 & 41.2 & 27.4 \\
        MVT~\cite{mvt} & 77.7 & 66.4 & 31.9 & 25.3 & 40.8 & 33.3 \\
        3DVG-Transformer~\cite{3dvg-trans} & 81.9 & 60.6 & 39.3 & 28.4 & 47.6 & 34.7 \\
        ViL3DRel~\cite{vil3drel} & 81.6 & 68.6 & 40.3 & 30.7 & 47.9 & 37.7 \\
        3DJCG~\cite{3djcg} & 83.5 & 64.3 & 41.4 & 30.8 & 49.6 & 37.3 \\
        D3Net~\cite{d3net} & -- & 72.0 & -- & 30.1 & -- & 37.9 \\
        M3DRef-CLIP~\cite{multi3drefer} & 85.3 & 77.2 & 43.8 & 36.8 & 51.9 & 44.7 \\
        3D-VisTA~\cite{3dvista} & 81.6 & 75.1 & 43.7 & 39.1 & 50.6 & 45.8 \\
        3D-LLM (Flamingo) \cite{3d-llm} & -- & -- & -- & -- & 21.2 & -- \\
        3D-LLM (BLIP2-flant5) \cite{3d-llm} & -- & -- & -- & -- & 30.3 & -- \\
        Grounded 3D-LLM \cite{grounded-3dllm} & -- & -- & -- & -- & 47.9 & 44.1 \\
        PQ3D \cite{pq3d} & 86.7 & 78.3 & 51.5 & 46.2 & 57.0 & 51.2 \\
        ChatScene \cite{chatscene} & {89.6} & {82.5} & 47.8 & 42.9 & 55.5 & 50.2 \\
        LLaVA-3D \cite{llava3d} & -- & -- & -- & -- & 54.1 & 42.2\\
        Video-3D LLM~\cite{video3dllm} & 88.0 & 78.3 & 50.9 & 45.3 & 58.1 & 51.7  \\
        
        \midrule

        baseline~\cite{video3dllm} &  82.17 & 73.71 & 45.10 & 40.14 & 52.29 & 46.66\\

        3DRS*~\cite{3drs} & 82.76 & 73.50  & 50.74 & 45.37 & 56.95  & 50.83\\
        
        \rowcolor{myblue}
        \textbf{3D-IDE (Ours)} & \textbf{86.72} & \textbf{77.94}  & \textbf{54.73} & \textbf{48.90} &  \textbf{60.94}  & \textbf{54.53} \\
        \bottomrule
    \end{tabular}
    }
    }
    \label{tab:scanrefer_detail}
\end{table*}

%% file: tabs/multirefer_detail.tex
\begin{table*}[hb]
    \centering
    \caption{\textbf{Performance comparison on Multi3DRefer validation set.} ZT: zero-target, ST: single-target, MT: multi-target, D: distractor.}
    \small
    \resizebox{\textwidth}{!}{
    \begin{tabular}{lcccccccccc}
        \toprule
        \multirow{2}{*}{Method}  & ZT w/o D & ZT w/ D & \multicolumn{2}{c}{ST w/o D} & \multicolumn{2}{c}{ST w/ D} &\multicolumn{2}{c}{MT} & \multicolumn{2}{c}{ALL} \\
         & F1 & F1 & F1@0.25 & F1@0.5 & F1@0.25 & F1@0.5 & F1@0.25 & F1@0.5 & F1@0.25 & F1@0.5 \\
         \midrule
        M3DRef-CLIP~\cite{multi3drefer} & 81.8 & 39.4 & 53.5 & 47.8 & 34.6 & 30.6 & 43.6 & 37.9 & 42.8 & 38.4 \\
        D3Net~\cite{d3net}  & 81.6 & 32.5 & -- & 38.6 & -- & 23.3 & -- & 35.0 & -- & 32.2 \\
        3DJCG~\cite{3djcg} & 94.1 & 66.9 & -- & 26.0 & -- & 16.7 & -- & 26.2 & -- & 26.6 \\
        Grounded 3D-LLM \cite{grounded-3dllm} & -- & -- & -- & -- & -- & -- & -- & -- & 45.2 & 40.6 \\
        PQ3D \cite{pq3d} & 85.4 & 57.7 & -- & 68.5 & -- & 43.6 & -- & 40.9 & -- & 50.1 \\
        ChatScene \cite{chatscene}  & 90.3 & 62.6 & {82.9} & {75.9} & 49.1 & 44.5 & {45.7} & {41.1} & 57.1 & 52.4 \\
        
        Video-3D LLM~\cite{video3dllm} & 94.7 & 78.5 & 82.6 & 73.4 & 52.1 & 47.2 & 40.8 & 35.7 & 58.0 & 52.7 \\
        3DRS ~\cite{3drs} & {95.6} & {79.4} & 79.6 & 71.4 & {57.0} & {51.3} & 43.0 & 37.8 & {60.4} & {54.9} \\
        
        \midrule

       baseline~\cite{video3dllm} & 98.7 & 91.5  & 60.5 & 54.9  & 36.9  & 33.8 & 35.9  &31.6 & 45.9 & 42.3\\

        3DRS*~\cite{3drs} & 96.6  & 85.2  & 75.1  & 67.4  & 49.0  & 44.8  & 42.6  & 37.6 & 55.8 & 51.1\\
     
        \rowcolor{myblue}
        \textbf{3D-IDE (Ours)} & {95.6} & {79.7} & \textbf{79.9} & \textbf{72.6} & \textbf{54.7} & \textbf{49.7} & \textbf{45.3} & \textbf{40.5} & \textbf{59.8} & \textbf{54.9} \\
        \bottomrule
    \end{tabular}
    }
    \label{tab:multi3drefer_detail}
\end{table*}

%% file: tabs/scanqa_detail.tex
\begin{table*}[h]
    \centering
    \caption{ \textbf{Performance comparison on the validation set of ScanQA.} EM indicates exact match accuracy, and B-1, B-2, B-3, B-4 denote BLEU-1, -2, -3, -4, respectively.}
    {\small
    \resizebox{\textwidth}{!}{
    \begin{tabular}{lcccccccc}
        \toprule
        Method & EM & B-1 & B-2 & B-3 & B-4 & ROUGE-L & METEOR & CIDEr \\
        \midrule
        ScanQA~\cite{scanqa} & 21.05 & 30.24 & 20.40 & 15.11 & 10.08 & 33.33 & 13.14 & 64.86  \\
        3D-VisTA \cite{3dvista} & 22.40 & -- & -- & -- & 10.40 & 35.70 & 13.90 & 69.60 \\
        Oryx-34B \cite{oryx} & -- & 38.00 & 24.60 & -- & -- & 37.30 & 15.00 & 72.30 \\
        LLaVA-Video-7B \cite{llavavideo} & -- & 39.71 & 26.57 & 9.33 & 3.09 & 44.62 & 17.72 & 88.70 \\
        3D-LLM (Flamingo) \cite{3d-llm} & 20.40 & 30.30 & 17.80 & 12.00 & 7.20 & 32.30 & 12.20 & 59.20 \\
        3D-LLM (BLIP2-flant5)~\cite{3d-llm} & 20.50 & 39.30 & 25.20 & 18.40 & 12.00 & 35.70 & 14.50 & 69.40  \\
        Chat-3D \cite{chat3d} & -- & 29.10 & -- & -- & 6.40 & 28.50 & 11.90 & 53.20 \\
        NaviLLM \cite{navillm} & 23.00 & -- & -- & -- & 12.50 & 38.40 & 15.40 & 75.90 \\
        LL3DA~\cite{ll3da} & -- & -- & -- & -- & 13.53 & 37.31 & 15.88 & 76.79 \\
        Scene-LLM~\cite{scenellm} & 27.20 & 43.60 & 26.80 & 19.10 & 12.00 & 40.00 & 16.60 & 80.00 \\
        LEO~\cite{leo} & -- & -- & -- & -- & 11.50 & 39.30 & 16.20 & 80.00 \\
        Grounded 3D-LLM \cite{grounded-3dllm} & -- & -- & -- & -- & 13.40 & -- & -- & 72.70\\
        ChatScene \cite{chatscene} & 21.62 & 43.20 & 29.06 & 20.57 & 14.31 & 41.56 & 18.00 & 87.70 \\
        LLaVA-3D \cite{llava3d} & 27.00 & -- & -- & -- & 14.50 & {50.10} & {20.70} &  91.70 \\
        Video 3D-LLM~\cite{video3dllm} & {30.10} & {47.05} & {31.70} & {22.83} & 16.17 & 49.02 & 19.84 & {102.0} \\

        \midrule

        baseline~\cite{video3dllm} & 29.5 & 46.9  & 31.3 & 22.7  & 16.2  & 48.8 & 19.6  &100.5\\

        3DRS*~\cite{3drs} & 29.7  &\textbf{47.9}  & 32.5  & \textbf{23.8} & 16.9  & 48.3 & 20.2  & 101.3 \\

        \rowcolor{myblue}
        \textbf{3D-IDE (Ours)} & \textbf{29.8} & {47.5} & \textbf{32.9} & {23.7} & \textbf{17.4} & \textbf{48.8} & \textbf{20.8} & \textbf{102.1} \\
        \bottomrule
    \end{tabular}
    }
    }
    \label{tab:scanqa_detail}
\end{table*}